\documentclass[10pt]{article}

\usepackage{mcr}
\usepackage{pkg}
\usepackage{algE}

\usepackage{natbib}

\newtheorem{lemma}{Lemma}

\DeclareMathOperator*{\argmin}{arg\,min}
\DeclareMathOperator*{\argmax}{arg\,max}

\title{
More Powerful and General Selective Inference for Stepwise Feature Selection Using the Homotopy Method}

\date{\today}

\makeatletter
\def\@fnsymbol#1{\ensuremath{\ifcase#1\or
{1}\or
{\dagger}\or
{2}\or
{*}\or 
\else\@ctrerr\fi}}
\makeatother

\author{
Kazuya Sugiyama\thanks{Nagoya Institute of Technology}  \thanks{Equal contribution} ,
Vo Nguyen Le Duy\footnotemark[1] \footnotemark[2] ,
Ichiro Takeuchi\footnotemark[1] \thanks{RIKEN} \thanks{Correspondence to: Ichiro Takeuchi (takeuchi.ichiro@nitech.ac.jp)} 
}

\begin{document}

\maketitle

\begin{abstract}
\noindent
Conditional selective inference (SI) has been actively studied as a new statistical inference framework for data-driven hypotheses.
The basic idea of conditional SI is to make inferences conditional on the selection event characterized by a set of linear and/or quadratic inequalities.
Conditional SI has been mainly studied in the context of feature selection such as stepwise feature selection (SFS).
The main limitation of the existing conditional SI methods is the loss of power due to over-conditioning, which is required for computational tractability.
In this study, we develop a more powerful and general conditional SI method for SFS using the homotopy method which enables us to overcome this limitation. 
The homotopy-based SI is especially effective for more complicated feature selection algorithms.
As an example, we develop a conditional SI method for forward-backward SFS with AIC-based stopping criteria, and show that it is not adversely affected by the increased complexity of the algorithm.
We conduct several experiments to demonstrate the effectiveness and efficiency of the proposed method.

\end{abstract}

\section{Introduction}
As machine learning (ML) is being applied to a greater variety of practical problems, ensuring the reliability of ML is becoming increasingly important.
Among several potential approaches to reliable ML, \emph{conditional selective inference} (SI) is recognized as a promising approach for evaluating the statistical reliability of data-driven hypotheses selected by ML algorithms.
The basic idea of conditional SI is to make inference on a data-driven hypothesis conditional on the selection event that the hypothesis is selected.
Conditional SI has been actively studied especially in the context of feature selection. 
Notably, \citet{lee2016exact} and \citet{tibshirani2016exact} proposed conditional SI methods for selected features using Lasso and stepwise feature selection (SFS), respectively. 
Their basic idea is to characterize the selection event by a polytope, i.e., a set of linear inequalities, in a sample space.
When a selection event can be characterized by a polytope, practical computational methods developed by these authors can be used for making inferences of the selected hypotheses conditional on the selection events.

Unfortunately, however, such \emph{polytope-based SI} has several limitations because it can only be used when the characterization of all relevant selection events is represented by a polytope.
In fact, in most of the existing polytope-based SI studies, \emph{extra-conditioning} is required in order for the selection event to be characterized as a polytope. 
For example, in SI for SFS by \citet{tibshirani2016exact}, the authors needed to consider conditioning not only on selected features but also on additional events regarding the \emph{history} of the feature selection algorithmic process and the \emph{signs} of the features. 
Such extra-conditioning leads to loss of power in the inference~\citep{fithian2014optimal}.

In this study, we go beyond polytope-based SI and propose a novel conditional SI method using the \emph{homotopy continuation} approach for SFS.
We call the proposed method $\emph{homotopy-based SI}$. 
In contrast to polytope-based SI for SFS in \citet{tibshirani2016exact}, the proposed method is more powerful and more general.
The basic idea of homotopy-based SI is to use the homotopy continuation approach to keep track of the hypothesis selection event when the dataset changes in the direction of the selected test statistic, which enables efficient identification of the subspace of the sample space in which the same hypothesis is selected. 
Furthermore, we show that the proposed homotopy-based SI is more advantageous for more complicated feature selection algorithms.
As an example, we develop a homotopy-based conditional SI for forward-backward SFS (FB-SFS) algorithm with AIC-based stopping criteria, and show that it is not adversely affected by the increased complexity of the algorithm and that a sufficiently high power is retained.
\paragraph{Related works}
Traditional statistical inference presumes that a statistical model and a statistical target for which we seek to conduct inference are determined before observing the dataset.
Therefore, if we apply traditional methods to hypotheses selected after observing the dataset, the inferential results are no longer valid.
This problem has been extensively discussed in the context of feature selection.
In fact, even in commonly used feature selection methods such as SFS, correct assessment of the statistical reliability of selected features has long been difficult.
Several approaches have been suggested in the literature toward addressing this problem~\citep{benjamini2005false,leeb2005model,leeb2006can,benjamini2009selective,potscher2010confidence,berk2013valid,lockhart2014significance,taylor2014post}.
A particularly notable approach is \emph{conditional} SI introduced in the seminal paper by \citet{lee2016exact}.
In their work, the authors showed that, for a set of features selected by Lasso, the selection event can be characterized as a polytope by conditioning on the selected set of features as well as additional events on their signs.
Furthermore, \citet{tibshirani2016exact} showed that polytope-based SI is also applicable to SFS by additionally conditioning on the \emph{history} and \emph{signs} of sequential feature selection.
Conditional SI has been actively studied in the past few years and extended to various directions~\citep{fithian2015selective,choi2017selecting,tian2018selective,chen2019valid,hyun2018post,loftus2014significance,loftus2015selective,panigrahi2016bayesian,tibshirani2016exact,yang2016selective,suzumura2017selective,tanizaki2019computing,duy2020computing}.

In conditional SIs, it is typically preferred to condition on as little as possible so that the inference can be more powerful \citep{fithian2014optimal}.
Namely, the main limitations of current polytope-based SI methods are that excessive conditioning is required to represent the selection event with a single polytope.
In the seminal paper by \citet{lee2016exact}, the authors already discussed the problem of over-conditioning, and explained how extra conditioning on signs can be omitted by an exhaustive enumeration of all possible signs and by taking the union over the resulting polyhedra.
However, such an exhaustive enumeration of exponentially increasing number of sign combinations is feasible only when the number of selected features is fairly small.
Several other approaches were proposed to circumvent the drawbacks and restrictions of polytope-based SI. 
\citet{loftus2015selective} extended polytope-based SI such that selection events characterized by quadratic inequalities can be handled, but this inherits the same over-conditioning problem.
To improve the power, \citet{tian2018selective} proposed an approach to randomize the algorithm in order to condition on less.
\citet{terada2019selective} proposed to use bootstrap re-sampling to characterize the selection event more generally.
The disadvantage of these approaches is that additional randomness is introduced into the algorithm and/or the inference.

This study is motivated by \citet{liu2018more} and \citet{duy2020parametric}.
The former studied Lasso SI for full-model parameters, whereas the latter extended the basic idea of the former so that it can be also applied to Lasso SI in more general settings.
These two studies go beyond polytope-based SI for more powerful Lasso SI without conditioning on signs.
Because Lasso is formulated as a convex optimization problem, the selection event for conditional SI can be characterized based on the optimality conditions of the convex optimization problem.
These two methods rely on the optimality conditions to characterize the minimum conditions for the Lasso feature selection event.

In contrast, the SFS algorithm cannot be formulated as an optimization problem.
Therefore, existing conditional SI methods for SFS are based on conditioning on the \emph{history} rather than the \emph{output} of the SFS algorithm.
Namely, existing SI methods actually consider conditions not only on which features are selected but also in what order they are selected.
Moreover, when the SFS algorithm is extended to more complex procedures such as FB-SFS, the over-conditioning problem has more pronounced negative effects.
Because features can be added and removed in various orders in FB-SFS, if we condition on the history rather than the output of the algorithm, we will suffer from extensive over-conditioning on the entire history of feature addition and removal.

\paragraph{Our contributions}
Our contributions are as follows.
First, we propose a novel conditional SI method for the SFS algorithm using the homotopy continuation approach and show that the proposed method overcomes the over-conditioning problem in existing methods and enables us to conduct minimally conditioned more powerful conditional SI on the selected features by the SFS algorithm.
Second, we develop a conditional SI method for the FB-SFS algorithm with AIC-based stopping criteria by using the homotopy continuation approach. 
Although the conventional polytope-based SI method is more adversely affected by the over-conditioning problem when SFS is extended to FB-SFS, we show that our homotopy-based method still enables us to conduct minimally-conditioned SI and retains sufficiently high power.
We demonstrate the validity and power of the proposed homotopy-based SI methods through numerical experiments on synthetic and real benchmark datasets. 
%
%
The codes will be released when the submission of this draft is accepted. 

\paragraph{Notations}
For a natural number $n$, we denote $[n]$ to be the set $\{1, \ldots, n\}$.
The $n \times n$ identity matrix is denoted by $I_n$.
For a matrix $X$ and a set of column indices $\cM$, $X_\cM$ indicates the matrix with the set of columns corresponding to $\cM$.

\section{Problem Statement}
\label{sec:problem_statement}
We consider the forward SFS for a regression problem.
Let $n$ be the number of instances and $p$ be the number of original features.
We denote the observed dataset as $(X, \bm y)$ where $X \in \RR^{n \times p}$ is the design matrix and $\bm y \in \RR^n$ is the response vector. 
Following the problem setup from the existing literature on conditional SI such as \citet{lee2016exact} and \citet{tibshirani2016exact}, we assume that the observed response $\bm y$ is a realization of the following random response vector
\begin{equation}
 \label{eq:random_vector}
 {\bm Y} = (Y_1, ..., Y_n)^\top \sim \NN({\bm \mu}, \Sigma),
\end{equation}
where ${\bm \mu} \in \RR^n$ is the unknown mean vector and $\Sigma \in \RR^{n \times n}$ is the covariance matrix, which is known or estimable from independent data.
The design matrix $X$ is assumed to be non-random.
For notational simplicity, we assume that each column vector of $X$ is normalized to have unit length.

\paragraph{Stepwise feature selection}
We consider the standard forward SFS method as studied in \citet{tibshirani2016exact} in \S\ref{sec:problem_statement} and \S\ref{sec:proposed_method}.
At each step of the SFS method, the feature that most improves the fit is newly added.
When each feature has unit length, it is equivalent to the feature which is most correlated with the residual of the least-square regression model fitted with the previously selected features.
For a response vector $\bm y \in \RR^n$ and a set of features $\cM \subseteq [p]$, let $\bm r(\bm y, X_\cM)$ be the residual vector obtained by regressing $\bm y$ onto $X_\cM$ for a set of features $\cM$, i.e., 
\begin{align*}
 \bm r(\bm y, X_\cM) = P_{X_\cM}^\perp \bm y = (I_n -  P_{X_{\cM}} ) \bm y
\end{align*}
where $P_{X_\cM} = X_\cM(X_\cM^\top X_\cM)^{-1}X_\cM^\top$.
Let $K$ be the number of the selected features by the SFS method\footnote{
We discuss a situation where the number of features $K$ is selected by cross-validation in Appendix \ref{app:cv}.
In other parts of the paper, we assume that $K$ is determined before looking at the data.
}.
We denote the index of the feature selected at step $k$ as $j_k$ and the set of selected features up to step $k$ as $\cM_k = \{j_1, \ldots, j_k\}$ for $k \in [K]$ (Note, however, that, if there is no ambiguity, we simply denote the final set of the selected features as $\cM~( = \cM_K)$).
Feature $j_k$ is selected as 
\begin{align}
 \nonumber
 j_k
 &
 = \argmin_{j \in [p] \setminus \cM_{k-1}} \left\| \bm r(\bm y, X_{\cM_{k-1} \cup \{j\}}) \right\|_2^2
 \\
 \label{eq:sfs_criterion}
 &
 = \argmax_{j \in [p] \setminus \cM_{k-1}} \left| \bm x_j^\top \bm r(\bm y, X_{\cM_{k-1}}) \right|, 
\end{align}
where $\cM_0(\bm y) := \emptyset$ and $\bm r(\bm y, \cM_0) := \bm y$. 

\paragraph{Statistical inference}
In order to quantify the statistical significance of the relation between the selected features and the response, we consider a statistical test for each coefficient of the selected model parameters 
\begin{align*}
 \hat{\bm \beta}_{\cM} = (X_{\cM}^\top X_{\cM})^{-1} X_{\cM}^\top \bm y.
\end{align*}
Note that the $j^{\rm th}$ coefficient is written as $\hat{\beta}_{\cM, j} = \bm \eta^\top \bm y$ by defining 
\begin{align}
 \label{eq:eta}
 \bm \eta = X_{\cM} \left( X^\top_{\cM} X_{\cM}\right)^{-1} \bm{e}_j,
\end{align} 
where $\bm{e}_j \in \RR^{|\cM|}$ is a unit vector, $j^{\rm th}$ element of which is 1 and 0 otherwise.
Note that $\bm \eta$ depends on $\cM$ and $j$, but we omit the dependence for notational simplicity. 
We consider the following statistical test for each coefficient $\beta_{\cM, j} = \bm \eta^\top \bm \mu$
\begin{align}
 \label{eq:hypotheses}
 {\rm H}_{0}: \beta_{\cM, j} = 0 ~~~ \text{vs.} ~~~ {\rm H}_{1}: \beta_{\cM, j} \neq 0,
\end{align}
where $\beta_{\cM, j}$ is the $j^{\rm th}$ element of the population least squares $\bm \beta_{\cM} = P_{\cM} \bm \mu$, i.e., the projection of $\bm \mu$ onto the column space of $X_{\cM}$.
Throughout the paper, we do not assume that the linear model is correctly specified, i.e., it is possible that $\bm \mu \neq X_{\cM} \bm \beta$ for any $\bm \beta \in \RR^{p}$. 
Even when the linear model is not correctly specified, $\bm \beta_{\cM}$ is still a well-defined best linear approximation. 

\paragraph{Conditional Selective Inference}
Because the target of the inference is selected by observing the data $\bm y$, if we naively apply a traditional statistical test to the problem in \eq{eq:hypotheses} as if the inference target is pre-determined, the result will be invalid (type-I error cannot be controlled at the desired significance level) owing to \emph{selection bias}. 
To address the selection bias problem, we consider \emph{conditional SI} introduced in \citet{lee2016exact} and \citet{tibshirani2016exact}. 
Let us write the SFS algorithm as the following function: 
\begin{align*}
 \cA: \bm y \mapsto \cM, 
\end{align*}
which maps a response vector $\bm y \in \RR^n$ to the set of the selected feature $\cM$ by the SFS algorithm. 

In conditional SI, the inference is conducted based on the following conditional sampling distribution of the test-statistic: 
\begin{equation}
 \label{eq:condition_model}
 \bm{\eta}^\top \bm Y \mid \left \{ \cA(\bm Y) = \cA(\bm y), {\bm q}({\bm Y}) = {\bm q} ({\bm y})\right \},
\end{equation}
where
\begin{align*}
 {\bm q}({\bm Y}) = (I_n - {\bm c} {\bm \eta}^\top) {\bm Y} \text{ with } \bm c = \Sigma {\bm \eta} ({\bm \eta}^\top \Sigma {\bm \eta})^{-1}
\end{align*}
is the nuisance parameter, which is independent of the test statistic.
The first condition $\cA(\bm Y) = \cA({\bm y})$ in \eq{eq:condition_model} indicates the event that the set of the selected features by the $K$-step SFS method with a random response vector $\bm Y$ is $\cM$, i.e., the same as those selected with the observed response vector $\bm y$.
The second condition ${\bm q}({\bm Y}) = {\bm q}({\bm y})$ indicates that the nuisance parameter for a random response vector $\bm Y$ is the same as that for the observed vector $\bm y$\footnote{
The ${\bm q}(\bm Y)$ corresponds to the component $\bm z$ in the seminal paper (see \citet{lee2016exact}, \S5, Eq. 5.2, and Theorem 5.2).
}.

To conduct the conditional inference for \eq{eq:condition_model}, the main task is to identify the conditional data space 
\begin{align}
 \label{eq:conditional_data_space}
 \cY = \{\bm Y \in \RR^n \mid \cA(\bm Y) = \cA({\bm y}), {\bm q}({\bm Y}) = {\bm q}({\bm y})\}.
\end{align}
Once $\cY$ is identified, we can easily compute the pivotal quantity
\begin{equation}
 \label{eq:pivotal_quantity}
 F^{\cZ}_{\bm{\eta}^\top \bm \mu, {\bm \eta}^\top \Sigma {\bm \eta}} (\bm{\eta}^\top \bm y) \mid \bm y \in \cY,
\end{equation}
where $F_{m, s^2}^\cZ$ is the c.d.f. of the truncated Normal distribution with mean $m$, variance $s^2$, and truncation region $\cZ$.
Later, we will explain how $\cZ$ is defined in \eq{eq:pivotal_quantity}.
The pivotal quantity is crucial for calculating $p$-value or obtaining confidence interval.
Based on the pivotal quantity, we can obtain \emph{selective $p$-value} \citep{fithian2014optimal} in the form of 
 \begin{align}
 \label{eq:selective_p_value}
 p^{\rm selective}_{j} = 2\ \min\{\pi_{j}, 1 - \pi_{j}\},
\end{align}
where $\pi_{j} = 1 - F^{\cZ}_{0, {\bm \eta}^\top \Sigma {\bm \eta}} (\bm{\eta}^\top {\bm y})$.
Furthermore, to obtain $1 - \alpha$ confidence interval for any $\alpha \in [0, 1]$, by inverting the pivotal quantity in (\ref{eq:pivotal_quantity}), we can find the smallest and largest values of $\bm{\eta}^\top \bm \mu$ such that the value of the pivotal quantity remains in the interval $\left[ \frac{\alpha}{2}, 1- \frac{\alpha}{2} \right]$ \citep{lee2016exact}. 

\paragraph{Characterization of the conditional data space $\cY$}
Using the second condition in \eq{eq:conditional_data_space}, the data in $\cY$ are restricted to a line (see \citet{liu2018more} and \citet{fithian2014optimal}).
Therefore, the set $\cY$ can be re-written, using a scalar parameter $z \in \RR$, as
\begin{equation}
 \label{eq:parametrized_data_space}
 \cY = \{\bm y(z) \in \RR^n \mid \bm y(z) = \bm a + \bm b z, z \in \cZ \}
\end{equation}
where ${\bm a} = {\bm q}(\bm y)$, ${\bm b} = \Sigma {\bm \eta} ({\bm \eta}^\top \Sigma {\bm \eta})^{-1} $, and 
\begin{equation}
 \label{eq:truncation_region_z}
 \cZ = \left \{ Z \in \RR \mid \cA({\bm y}(Z)) = \cA({\bm y}) \right \}.
\end{equation}
Here, $\bm a$, $\bm b$ and $\cZ$ depend on $\cM$ and $j$, but we omit the subscripts for notational simplicity. 
Now, let us consider a random variable $Z \in \RR$ and its observation $z \in \RR$ such that they respectively satisfy ${\bm Y} = {\bm a} + {\bm b} Z$ and ${\bm y} =  {\bm a} + {\bm b} z$. 
The conditional inference in (\ref{eq:condition_model}) is re-written as the problem of characterizing the sampling distribution of 
\begin{equation}
 \label{eq:condition_parametric}
 Z \mid Z \in \cZ.
\end{equation}
Because $Z \sim \NN(\bm \eta^\top \bm \mu, {\bm \eta}^\top \Sigma {\bm \eta})$, $Z \mid Z \in \cZ$ follows a truncated Normal distribution.
Once the truncation region $\cZ$ is identified, the pivotal quantity in (\ref{eq:pivotal_quantity}) is obtained as $F^{\cZ}_{\bm \eta^\top \bm \mu, {\bm \eta}^\top \Sigma {\bm \eta}} (z)$.
Thus, the remaining task is reduced to the characterization of $\cZ$.

\paragraph{Over-conditioning in existing conditional SI methods}
Unfortunately, full identification of the truncation region $\cZ$ in conditional SI for the SFS algorithm is considered computationally infeasible. 
Therefore, in existing conditional SI studies such as \citet{tibshirani2016exact}, the authors circumvent the computational difficulty by \emph{over conditioning}. 
Note that over-conditioning is not harmful for selective type-I error control, but it leads to the loss of power \citep{fithian2014optimal}.
In fact, the decrease in the power due to over-conditioning is not unique problem for SFS in \citet{tibshirani2016exact} but is a common major problem in many existing conditional SIs~\citep{lee2016exact}.
In the next section, we propose a method to overcome this difficulty.

\section{Proposed Homotopy-based SI for SFS}
\label{sec:proposed_method}

As we discussed in \S2, to conduct conditional SI, the truncation region $\cZ \subseteq \RR$ in \eq{eq:truncation_region_z} should be identified.
To construct $\cZ$, our idea is 1) computing $\cA({\bm y}(z))$ for all $z \in \RR$ and 2) identifying the set of intervals of $z \in \RR$ on which $\cA({\bm y}(z)) = \cA(\bm y)$. 
However, it seems intractable to obtain $\cA({\bm y}(z))$ for infinitely many values of $z \in \RR$.
To overcome the difficulty, we combine two approaches called \emph{extra-conditioning} and \emph{homotopy continuation}. 
Our idea is motivated by the regularization paths of Lasso \citep{osborne2000new, Efron04a}, SVM \citep{HasRosTibZhu04} and other similar methods \citep{RosZhu07, BacHecHor06, RosZhu07, Tsuda07, Lee07, Takeuchi09a, takeuchi2011target, Karasuyama11, hocking11a, Karasuyama12a, ogawa2013infinitesimal, takeuchi2013parametric}, in which the solution path along the regularization parameter can be computed by analyzing the KKT optimality conditions of parametrized convex optimization problems. 
Although SFS cannot be formulated as a convex optimization problem, by introducing the notion of extra-conditioning, we note that a conceptually similar approach as homotopy continuation can be used to keep track all possible changes in the selected features when the response vector $\bm y$ changes along the direction of the test-statistic.
A conceptually similar idea has recently been used for evaluating the reliability of deep learning representations \citep{duy2020quantifying}.


\subsection{Extra-Conditioning}
First, let us consider the \emph{history} $\cH$ and \emph{signs} $\cS$ of the SFS algorithm.
The \emph{history} of the SFS algorithm is defined as
\begin{align*}
 \cH := (\cM_1, \cM_2, \ldots, \cM_K), 
\end{align*}
i.e., the sequence of the sets of the selected features in $K$ steps.
The \emph{signs} of the SFS algorithm are defined as a vector $\cS := (\cS_1, \cS_2, \ldots, \cS_K)$, $k^{\rm th}$ element of which is defined as 
\begin{align*}
 \cS_k = {\rm sgn}
 \left(
 \bm x_{j_k}^\top \bm r(\bm y, X_{\cM_{k-1}})
 \right)
\end{align*}
for $k \in [K]$, which indicates the sign of the $j_k^{\rm th}$ feature when it is first entered to the model at step $k$.

We interpret the function $\cA: \bm y \mapsto M$ as a composite function $\cA = \cA_1 \circ \cA_2$  where
\begin{align*}
 \cA_1: (\cH, \cS) \mapsto \cM,
\text{ and }
 \cA_2: \bm y \mapsto (\cH, \cS),
\end{align*}
i.e.,
the following relationships hold:
$
\cM = \cA(\bm y) = \cA_1(\cA_2(\bm y)),
\cM = \cA_1((\cH, \cS)),
(\cH, \cS) = \cA_2(\bm y)
$.

Let us now consider conditional SI not on $\cA(\bm Y) = \cA(\bm y)$ but on $\cA_2(\bm Y) = \cA_2(\bm y)$. 
The next lemma indicates that, by conditioning on the history and the signs $(\cH, \cS)$ rather than the set of the selected features $\cM$, the truncation region can be simply represented as an interval in the line $\bm y(z) = \bm a + \bm b z, z \in \RR$. 
\begin{lemma}
 \label{lemm:over_condition}
 Consider a response vector $\bm y^\prime \in \cY$. 
 Let $(\cH^\prime, \cS^\prime) = \cA_2(\bm y^\prime)$ be the history and the signs obtained by applying the $K$-step SFS algorithm to the response vector $\bm y^\prime$, and their elements are written as 
 \begin{align*}
  \cH^\prime = (\cM_1^\prime, \ldots, \cM_K^\prime)
  \text{ and }
  \cS^\prime = (\cS_1^\prime, \ldots, \cS_K^\prime).
 \end{align*}
 Then, the over-conditioned truncation region defined as 
\begin{align}
 \label{eq:z_oc}
 \cZ^{\rm oc}(\bm y^\prime) = \left\{z \in \RR \mid \cA_2(\bm y(z)) = \cA_2(\bm y^\prime) \right\}
\end{align}
 is an interval
 \begin{align}
  \label{eq:lemm1_c}
  z
  \in
  \left[
  \max_{\substack{ k \in [K], \\ j \in [p] \setminus \cM^\prime_{k-1}, \\ d_{(k,j)} > 0} } \frac{e_{(k,j)}}{d_{(k,j)}}, 
  \min_{\substack{k \in [K], \\ j \in [p] \setminus \cM^\prime_{k-1}, \\ d_{(k,j)} < 0}} \frac{e_{(k,j)}}{d_{(k,j)}}
  \right],
 \end{align}
 where 
 \begin{align*}
  e_{(k,j)}
  &
  =(\bm x_{j} - \bm x_{j_k} \cS^\prime_{k})^\top P_{X_{\cM_{k-1}}}^\perp \bm a, 
  \\
  d_{(k,j)}
  &
  = (\bm x_{j_k} \cS^\prime_{k} - \bm x_{j} )^\top P_{X_{\cM^\prime_{k-1}}}^\perp \bm b,
 \end{align*}
 for $k \in [K], j \in [p] \setminus \cM^\prime_{k-1}$.
\end{lemma}
The proof is presented in Appendix \ref{app:proofs}. 

Note that the condition $\cA_2(\bm Y) = \cA_2(\bm y)$ is redundant for our goal of making the inference conditional on $\cA(\bm Y) = \cA(\bm y)$. 
This over-conditioning is undesirable because it decreases the power of conditional inference~\citep{fithian2014optimal}, but the original conditional SI method for SFS in \citet{tibshirani2016exact} performs conditional inference under exactly this over-conditioning case because otherwise the selection event cannot be formulated as a polytope. 
In the following subsection, we overcome this computational difficulty by utilizing the homotopy continuation approach.


\subsection{Homotopy Continuation}

In order to obtain the optimal truncation region $\cZ$ in \eq{eq:truncation_region_z}, our idea is to enumerate all the over-conditioned regions in \eq{eq:z_oc} and to consider their union 
\begin{align*}
 \cZ = \bigcup_{\bm y^\prime \in \cY \mid \cA(\bm y^\prime) = \cA(\bm y)} \cZ^{\rm oc}(\bm y^\prime)
\end{align*}
by homotopy continuation. 
To this end, let us consider all possible pairs of history and signs $(\cH(z), \cS(z)) = \cA_2(\bm y(z))$ for $z \in \RR$.
Clearly, the number of such pairs is finite.
With a slight abuse of notation, we index each pair by $t=1, 2, \ldots, T$ and denote it such as $(\cH^{(t)}, \cS^{(t)})$, where $T$ is the number of the pairs. 
Lemma 1 indicates that each pair corresponds to an interval in the line. 
Without loss of generality, we assume that the left-most interval corresponds to $t=1$, the second left-most interval corresponds to $t=2$, and so on. 
Then, using an increasing sequence of real numbers $z_1 < z_2 < \cdots < z_T < z_{T+1}$, we can write these intervals as $[z_1, z_2], [z_2, z_3], \ldots, [z_{T}, z_{T+1}]$.
In practice, we do not have to consider the entire line $z \in (-\infty, \infty)$, but it suffices to consider a sufficiently wide range.
Specifically, in the experiments in \S\ref{sec:experiment}, we set $z_{\rm min} = z_1 = - (|z| + 10) \sigma$ and $z_{\rm max} = z_{T+1} = (|z| + 10) \sigma$ where $\sigma$ is the standard error of the test-statistic.
With this choice of the range, the probability mass outside the range is negligibly small.

Our simple idea is to compute all possible pairs $\{(\cH^{(t)}, \cS^{(t)})\}_{t=1}^T$ by keeping track of the intervals $[z_1, z_2], [z_2, z_3], \cdots, [z_T, z_{T+1}]$ one by one, and then to compute the truncation region $\cZ$ by collecting the intervals in which the set of the selected features $\cM^{(t)} = \cA_1((\cH^{(t)}, \cS^{(t)}))$ is the same as the set of the actually selected features from the observed data $\cM = \cA(\bm y)$, i.e., 
\begin{align*}
 \cZ = \bigcup_{t \in [T] \mid \cA_1((\cH^{(t)}, \cS^{(t)})) = \cA(\bm y)} [z_t, z_{t+1}].
\end{align*}
We call $z_1, z_2, \ldots, z_{T+1}$ \emph{breakpoints}.
We start from applying the SFS algorithm to the response vector $\bm y(z_1)$ and obtain the first pair $(\cH^{(1)}, \cS^{(1)})$. 
Next, using Lemma \ref{lemm:over_condition} with $\bm y^\prime = \bm y(z_1)$, the next breakpoint $z_2$ is obtained as the right end of the interval in \eq{eq:lemm1_c}.
Next, the second pair $(\cH^{(2)}, \cS^{(2)})$ is obtained by applying the SFS method to the response vector $\bm y(z_2 + \Delta z)$, where $\Delta z$ is a small value such that $z_t + \Delta z < z_{t+1}$ for all $t \in [T]$.
This process is repeated until the next breakpoint becomes greater than $z_{\rm max}$. 
The pseudo-code of the entire method is presented in Algorithm \ref{alg:parametric_sfs_SI} and the computation of the truncation region $\cZ$ is described in Algorithm \ref{alg:solution_path}.

\begin{algorithm}[!t]
\renewcommand{\algorithmicrequire}{\textbf{Input:}}
\renewcommand{\algorithmicensure}{\textbf{Output:}}
\begin{footnotesize}
 \begin{algorithmic}[1]
  \REQUIRE $X, {\bm y}, K, [z_{\rm min}, z_{\rm max}]$
	\vspace{2pt}
	\STATE $\cM \leftarrow$ Applying the $K$-step SFS algorithm to $(X, {\bm y}^{\rm obs})$
	\vspace{2pt}
	\FOR {each selected feature $j \in \cM$}
		\vspace{2pt}
		\STATE Compute $\bm \eta$ $\leftarrow$ Equation (\ref{eq:eta}) 
		\vspace{2pt}
		\STATE Compute $\bm a$ and $\bm b$ $\leftarrow$ Equation (\ref{eq:parametrized_data_space})
		\vspace{2pt}
		\STATE $\cZ \leftarrow {\tt compute\_truncation\_region}$ ($X$, $K$, $\bm a$, $\bm b$, $[z_{\rm min}, z_{\rm max}], \cM$)
		\vspace{2pt}
		\STATE $p^{\rm selective}_{j} \leftarrow $ Equation (\ref{eq:selective_p_value}) (and$/$or selective confidence interval)
		\vspace{2pt}
	\ENDFOR
	\vspace{2pt}
  \ENSURE $\{p^{\rm selective}_{j}\}_{j \in \cM}$ (and$/$or selective confidence intervals)
 \end{algorithmic}
\end{footnotesize}
\caption{{\tt SFS\_conditional\_SI}}
\label{alg:parametric_sfs_SI}
\end{algorithm}

\begin{algorithm}[!t]
\renewcommand{\algorithmicrequire}{\textbf{Input:}}
\renewcommand{\algorithmicensure}{\textbf{Output:}}
\begin{footnotesize}
 \begin{algorithmic}[1]
  \REQUIRE $X, K, \bm a, \bm b, [z_{\rm min}, z_{\rm max}], \cM$
  \vspace{2pt}
  \STATE Initialization: $t = 1$, $z_t=z_{\rm min}$, $\cZ = \emptyset$
  \vspace{2pt}
  \WHILE {$z_t < z_{\rm max}$}
  \vspace{2pt}
  \STATE $\bm y(z_t + \Delta z) = \bm a + \bm b (z_t + \Delta z)$
  \vspace{2pt}
  \STATE $(\cH^{(t)}, \cS^{t}) \leftarrow$ Applying $K$-step SFS to $(X, \bm y (z_t + \Delta z))$
  \vspace{2pt}
  \STATE $z_{t + 1} \leftarrow $ Equation \eq{eq:lemm1_c}
  \vspace{2pt}
  \IF {$\cM^{(t)} = \cM$}
  \STATE $\cZ \leftarrow \cZ \cup [z_t, z_{t+1}]$
  \ENDIF
  \vspace{2pt}
  \STATE $t \leftarrow t+1$
  \vspace{2pt}  
  \ENDWHILE
  \vspace{2pt}
  \ENSURE $\cZ = \{z \in [z_{\rm min}, z_{\rm max}] \mid \cA(\bm y(z)) = \cA(\bm y)\}$
 \end{algorithmic}
\end{footnotesize}
\caption{{\tt compute\_truncation\_region}}
\label{alg:solution_path}
\end{algorithm}

\section{Forward Backward Stepwise Feature Selection}
\label{sec:fb_sfs}
In this section, we present a conditional SI method for the \emph{FB-SFS} algorithm using the homotopy method.
At each step of the FB-SFS algorithm, a feature is considered for addition to or subtraction from the current set of the selected features based on some pre-specified criterion. 
As an example of commonly used criterion, we study the AIC-based criterion, which is used as the default option in well-known {\tt stepAIC} package in R. 

Under the assumption of Normal linear model $\bm Y \sim \cN(X \bm \beta, \Sigma)$, for a set of the selected features $\cM$, the AIC is written as 
\begin{align}
 \nonumber
 {\rm AIC}
 :=
 &
 \max_{\bm \beta} (\bm y - X_\cM \bm \beta)^\top \Sigma^{-1} (\bm y - X_\cM \bm \beta) + 2|\cM|
 \\
 \label{eq:aic1}
 =
 &
 \bm y^\top A_\cM \bm y + 2|\cM|, 
\end{align}
where $A_\cM = \Sigma^{-1} - \Sigma^{-1} X_\cM (X_\cM^\top \Sigma^{-1} X_\cM)^{-1} X_\cM^\top \Sigma^{-1}$ and $|\cM|$ is the number of the selected features (the irrelevant constant term is omitted here).
The goal of AIC-based FB-SFS is to find a model with the smallest AIC while adding and removing features step by step. 
The algorithm starts either from the null model (the model with only a constant term) or the full model (the model with all the $p$ features). 
At each step, among all possible choices of adding one feature to or deleting one feature from the current model, the one with the smallest AIC is selected. 
The algorithm terminates when the AIC is no longer reduced.

With a slight abuse of notations, let us use the following notations, the same as in \S\ref{sec:proposed_method}.
\begin{align*}
 \cA: \bm y \mapsto \cM, \cA_1: \cH \mapsto \cM, \cA_2: \bm y \mapsto \cH, 
\end{align*}
where $\cM$ is the set of the selected features and $\cH$ is the history of the FB-SFS algorithm written as
\begin{align*}
 \cH = (\cM_1, \ldots, \cM_K), 
\end{align*}
where $\cM_k$ is the set of the selected features at step $k$ for $ k \in [K]$.
Here, the history $\cH$ contains the information on feature addition and removal in the FB-SFS algorithm. 
Therefore, unlike the forward SFS algorithm in \S\ref{sec:proposed_method}, the size of $\cM_k$ can be increased or decreased depending on whether a feature is added or removed.

Using these notations, the truncation region $\cZ$ is similarly obtained as 
\begin{align}
 \nonumber
 \cZ
 &
 =
 \left\{
 z \in \RR
 \mid
 \cA(\bm y(z)) = \cA(\bm y)
 \right\}
 \\
 \label{eq:z_quadratic_a2}
 &
 =
 \bigcup_{\cH \mid \cA_1(\cH) = \cA(\bm y)}
 \left\{
 z \in \RR
 \mid
 \cA_2(\bm y(z)) = \cH
 \right\}
\end{align}
The basic idea of the homotopy method is the same as before; we find a range of $z \in \RR$ such that the algorithm has a certain history $\cH$ and take the union of those for which the output $\cM$ obtained from the history $\cH$ is equal to the actual selected set of features $\cM = \cA(\bm y)$.
The following lemma indicates that a range of $z \in \RR$ in which the algorithm has a certain history $\cH$ can be analytically obtained.
\begin{lemma}
 \label{lemm:quadratic}
 Consider a response vector $\bm y^\prime \in \cY$.
 Then, the over-conditioned truncation region defined as
 \begin{align}
  \label{eq:over_conditioned_z2}
  \cZ^{\rm oc}(\bm y^\prime)
  =
  \left\{
  z \in \RR
  \mid
  \cA_2(\bm y(z)) = \cA_2(\bm y^\prime)
  \right\}
 \end{align}
 is characterized by a finite number of quadratic inequalities of $z \in \RR$. 
\end{lemma}
The proof is presented in Appendix \ref{app:proofs}. 

Unlike the case of polytope-based SI for forward SFS in the previous section, the over-conditioned region $\cZ^{\rm oc}(\bm y^\prime)$ in Lemma~\ref{lemm:quadratic} possibly consists of multiple intervals.
The homotopy continuation approach can be similarly used for computing the union of intervals in \eq{eq:z_quadratic_a2}. 
Starting from $z_1 = z_{\rm min}$, we first compute $\cZ^{\rm oc}(\bm y^\prime)$ with $\bm y^\prime = \bm y(z_1)$ using Lemma~\ref{lemm:quadratic}.
If $\cZ^{\rm oc}(\bm y^\prime)$ consists of multiple intervals, we can consider only the interval containing $\bm y^\prime$ and set the right end of that interval as the next breakpoint. 
The computation of the next breakpoint $z_t$ can be written as 
\begin{align}
 \label{eq:update_bp2}
 z_t = \max\{z \in \cZ^{\rm oc}(\bm y(z_{t-1} + \Delta z))\} + \Delta z, 
\end{align}
for $t = 1, 2, \ldots, T$, where, if $\cZ^{\rm oc}(\bm y(z_{t-1} + \Delta z))$ consists of two separated intervals, then the maximum operator in \eq{eq:update_bp2} shall take the maximum in the interval containing $z_{t-1}$.

\section{Experiment}
\label{sec:experiment}
We present the experimental results of forward SFS in \S\ref{sec:proposed_method} and \S\ref{sec:fb_sfs} in \ref{subsec:exp_f} and \ref{subsec:exp_fb}, respectively.
The details of the experimental setups are described in Appendix~\ref{app:exp_setup}.
More experimental results on the computational and robustness aspects of the proposed method are provided in Appendices \ref{app:exp_computational} and \ref{app:exp_robustness}, respectively. 
In all the experiments, we set the significance level $\alpha = 0.05$.

\subsection{Forward SFS}
\label{subsec:exp_f}

We compared the following five methods: 
\begin{itemize}
 \item {\tt Homotopy}: conditioning on the selected features $\cM$ (minimal conditioning);
 \item {\tt Homotopy-H}: additionally conditioning on the history $\cH$;
 \item {\tt Homotopy-S}: additionally conditioning on the signs $\cS$;
 \item {\tt Polytope} \citep{tibshirani2016exact}: additionally conditioning on the history $\cH$ and signs $\cS$;
 \item {\tt DS}: split the data and use one for feature selection and the other for inference.
\end{itemize}

\paragraph{Synthetic data experiments}
We examined the false positive rates (FPRs), true positive rates (TPRs) and confidence interval (CI) lengths.
We generated the dataset $\{(\bm x_i, y_i)\}_{i \in [n]}$ by $\bm x_i \sim \NN(0, I_p)$ and $y_i = \bm x_i^\top \bm \beta + \veps_i$ with $\veps_i \sim \NN(0, 1)$. 
We set $n \in \{50, 100, 150\}$, $p=5$, $K=3$ for experiments on FPRs and TPRs. 
The coefficients $\bm \beta$ was respectively set as $[0, 0, 0, 0, 0]^\top$ and $[0.25, 0.25, 0, 0, 0]^\top$ for FPR and TPR experiments. 
We set $n=100$, $p=10$, $K=9$ and $\bm \beta = [0.25, 0.25, 0.25, 0.25, 0.25, 0, 0, 0, 0, 0]^\top$ for experiments on CIs.

\begin{figure}[p]
 \begin{center}
  \includegraphics[width=0.50\linewidth]{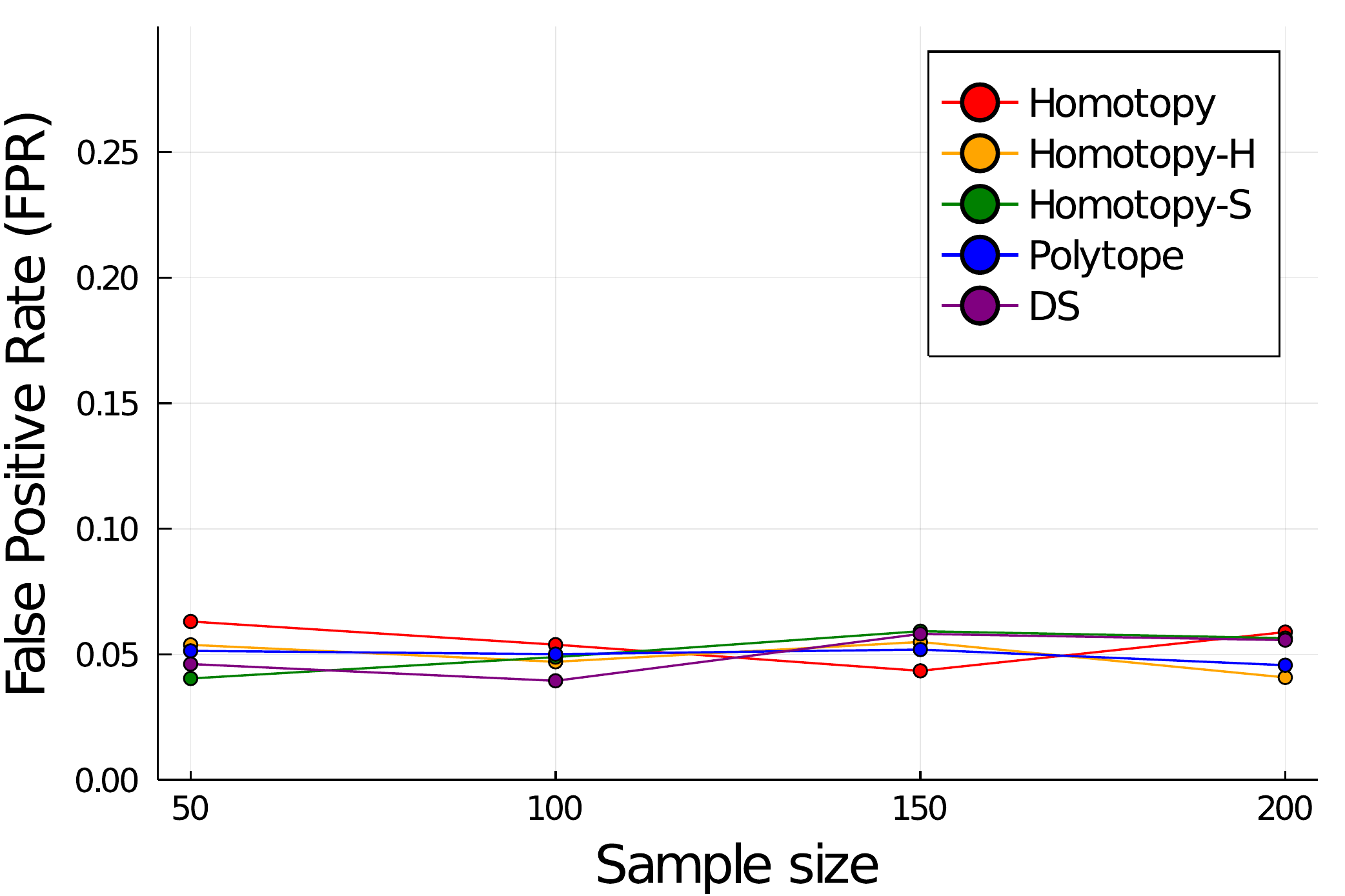}\\
  {\small (a) False Positive Rates (FPRs)} \\
  \includegraphics[width=0.50\linewidth]{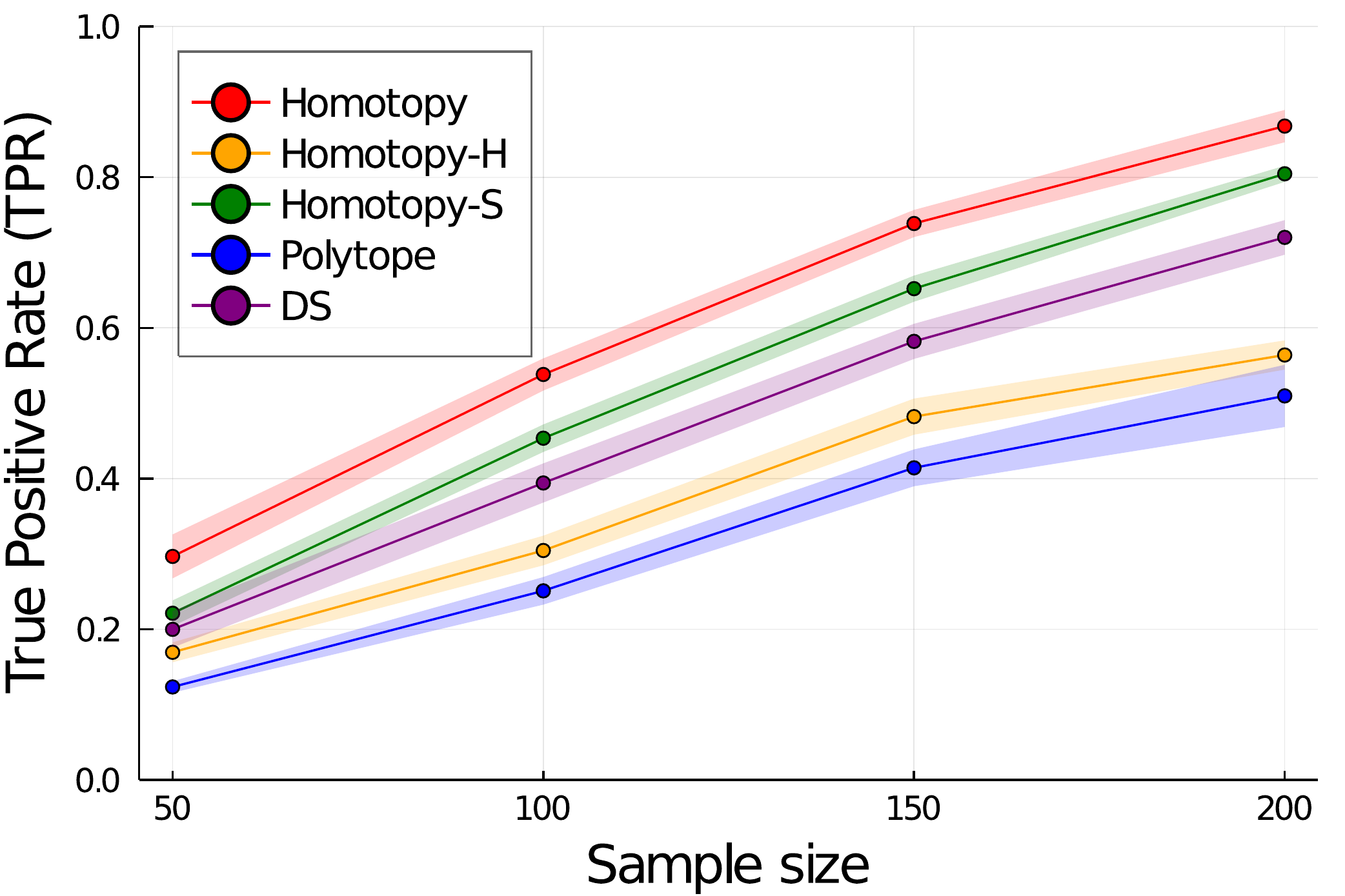}\\
  {\small (b) True Positive Rates (TPRs)} \\
  \includegraphics[width=0.50\linewidth]{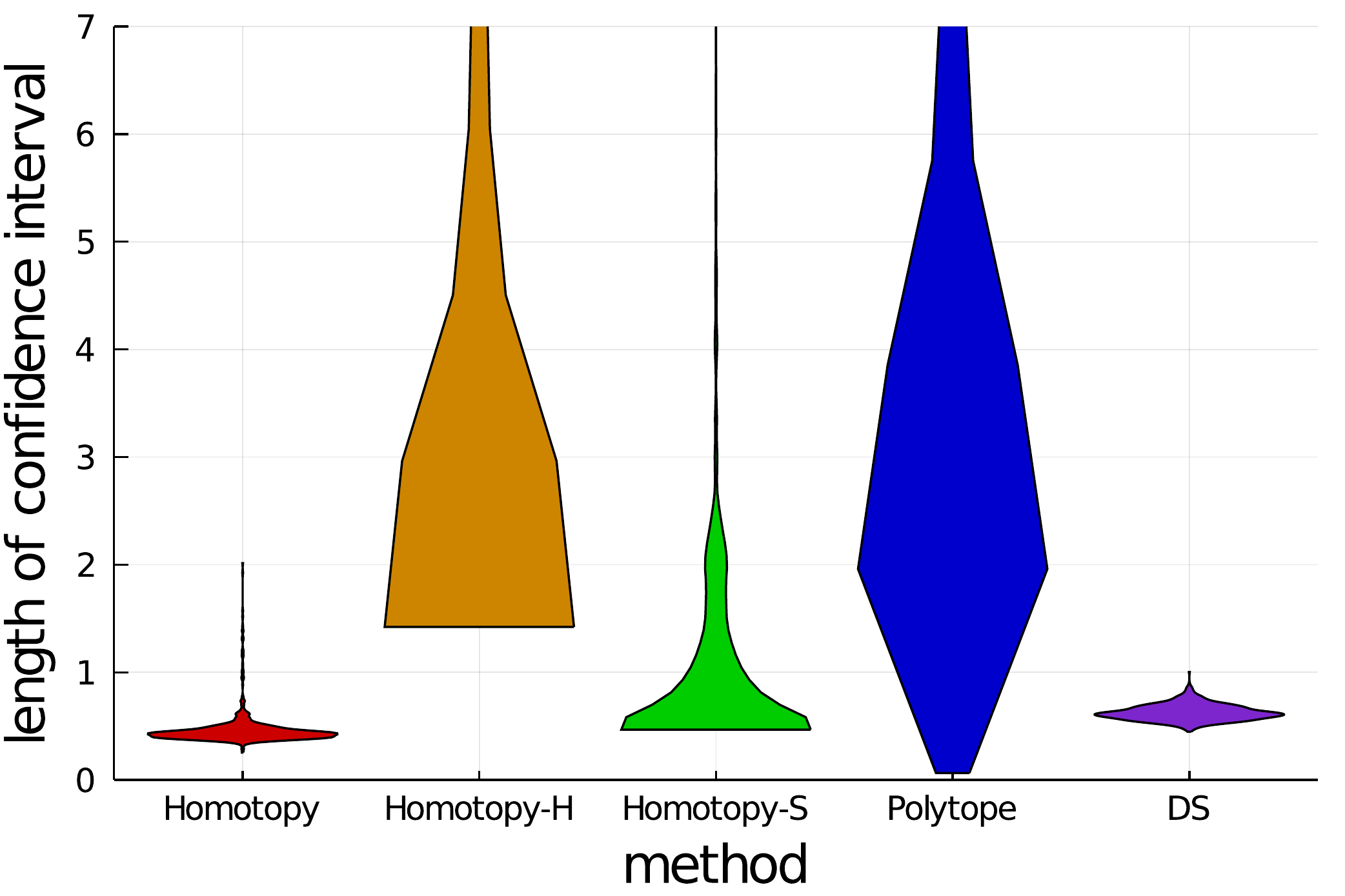}\\
  {\small (c) Confidence Interval (CI) Lengths} \\
  \caption{Results of forward SFS on synthetic data.}
  \label{fig:exp_f_sfs_synthetic}  
 \end{center}
\end{figure}

The results of FPRs, TPRs and CIs are shown in Fig. \ref{fig:exp_f_sfs_synthetic}(a), (b) and (c), respectively.
The FPRs and TPRs were estimated by 100 trials, and the plots in Fig. \ref{fig:exp_f_sfs_synthetic}(a) and (b) are the averages when the 100 trials were repeated 20 times, while the plots in Fig. \ref{fig:exp_f_sfs_synthetic}(c) are the results of 100 trials. 
In all five methods, the FPRs are properly controlled under the significance level $\alpha$.
Regarding the TPR comparison, it is obvious that {\tt Homotopy} method has the highest power because it is minimally conditioned. 
Note that the additional conditioning on the history $\cH$ and/or the signs $\cS$ significantly decreases the power.
The results of the CIs are consistent with the results of TPRs. 

\paragraph{Real data experiments}
We compared the proposed method ({\tt Homotopy}) and conventional method ({\tt Polytope}) on three real datasets\footnote{We used Housing (Dataset 1), Abalone (Dataset 2), and Concrete Compressive Strength (Dataset 3) datasets in UCI machine learning repository.}.
From each of the original dataset, we randomly generated sub-sampled datasets with sizes 25, 50 and 100. 
We applied the proposed and the conventional methods to each sub-sampled dataset, examined the cases where the selective $p$-values of a selected feature was different between the two methods, and computed the percentage of times when the proposed method provided smaller selective $p$-values than the conventional one. 
Table \ref{tab:exp_f_real} shows the results from 1000 trials. 
The proposed method provides smaller selective $p$-values than the conventional method significantly more frequently especially when $n$ is large.

\begin{table}[ht]
 \begin{center}
  \caption{Results on three real data experiments for forward SFS. Percentage of cases where selective $p$-values of the proposed method ({\tt Homotopy}) was smaller than that of the conventional method ({\tt Polytope}) in random sub-sampling experiments were shown.}
  \label{tab:exp_f_real}
  \begin{tabular}{l|c|c|c}
   & Dataset 1 & Dataset 2 & Dataset 3 \\ \hline
   $n =  25$ & 56.40\% & 57.69\% & 66.79\% \\
   $n =  50$ & 62.85\% & 65.1\% & 70.09\% \\
   $n = 100$ & 71.3\% & 71.88\% & 76.58\% 
  \end{tabular}
 \end{center}
\end{table}

\begin{table}[ht]
 \begin{center}
  \caption{Results on three real data experiments for FB-SFS. Percentage of cases where selective $p$-values of the proposed method ({\tt Homotopy}) was smaller than that of the conventional method ({\tt Polytope}) in random sub-sampling experiments were shown.}
  \label{tab:exp_fb_real}
  \begin{tabular}{l|c|c|c}
   & Dataset 1 & Dataset 2 & Dataset 3 \\ \hline
   $n =  25$ & 69.35\% & 75.43\% & 57.13\% \\
   $n =  50$ & 72.72\% & 78.42\% & 60.91\% \\
   $n = 100$ & 78.60\% & 81.38\% & 76.19\% 
  \end{tabular}
 \end{center}
\end{table}

\subsection{Forward-Backward (FB)-SFS}
\label{subsec:exp_fb}

We compared the following two methods: 
\begin{itemize}
 \item {\tt Homotopy}: conditioning on the selected features $\cM$ (minimal conditioning);
 \item {\tt Quadratic}: additionally conditioning on the history $\cH$ (implemented by using quadratic inequality-based conditional SI in \citep{loftus2015selective}).
\end{itemize}

\paragraph{Synthetic data experiments}
We examined the FPRs, TPRs and CI lengths by generating synthetic datasets in the same way as above.
We set $n \in \{50, 100, 150\}$ $p \in \{10, 20, 50\}$ for experiments on FPRs and TPRs. 
The coefficients $\bm \beta$ were zero for the case of FPR, whereas the first half of the coefficients were either $0.01, 0.25, 0.5, 1$ and the second half were zero for the case of TPR. 
We set $n=100$, $p=10$, and each element of $\bm \beta$ was randomly set from $\NN(0,1)$ for experiments on CIs.

\begin{figure}[p]
 \begin{center}
  \includegraphics[width=0.50\linewidth]{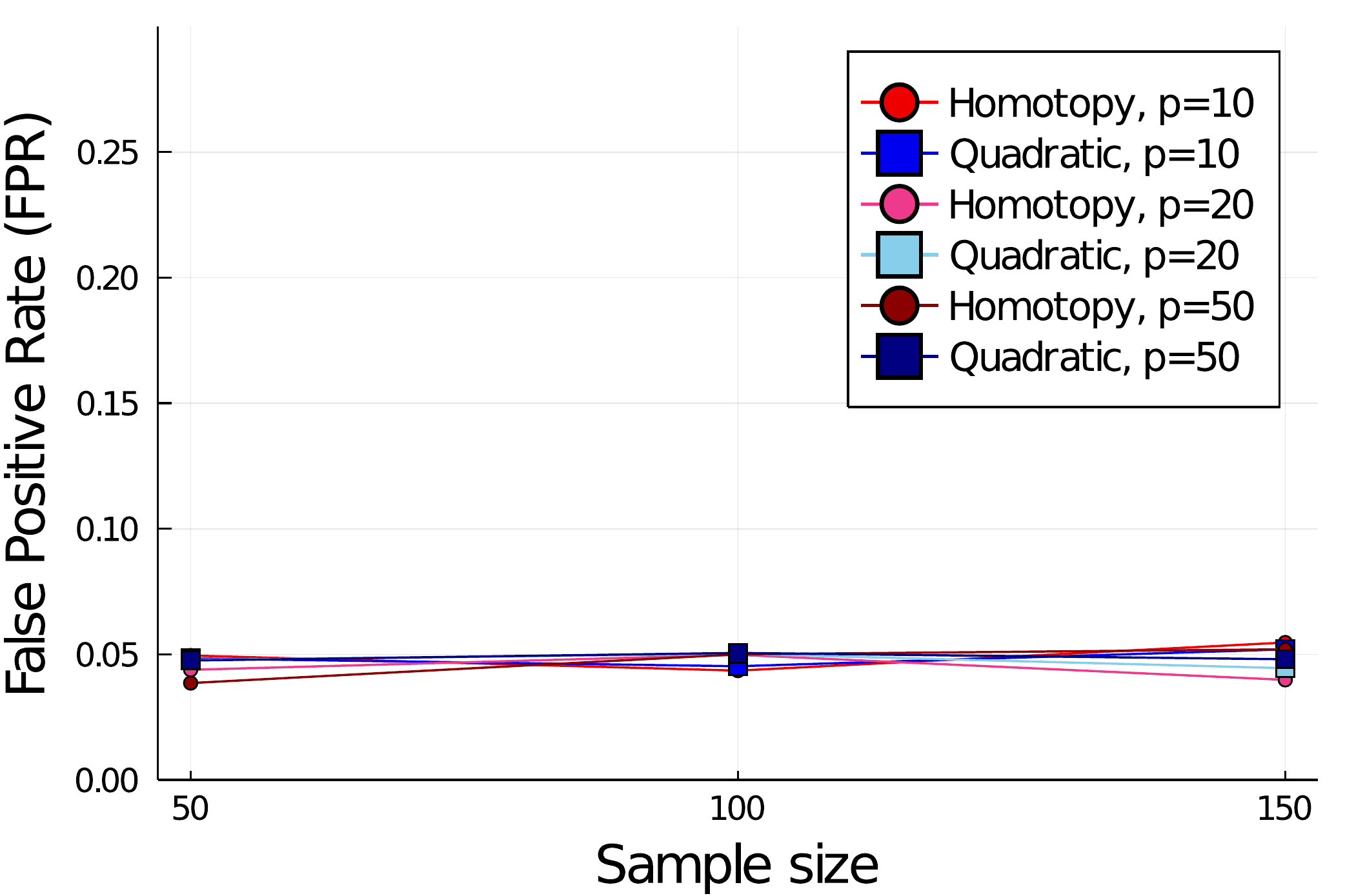}\\
  {\small (a) False Positive Rates (FPRs)} \\
  \includegraphics[width=0.50\linewidth]{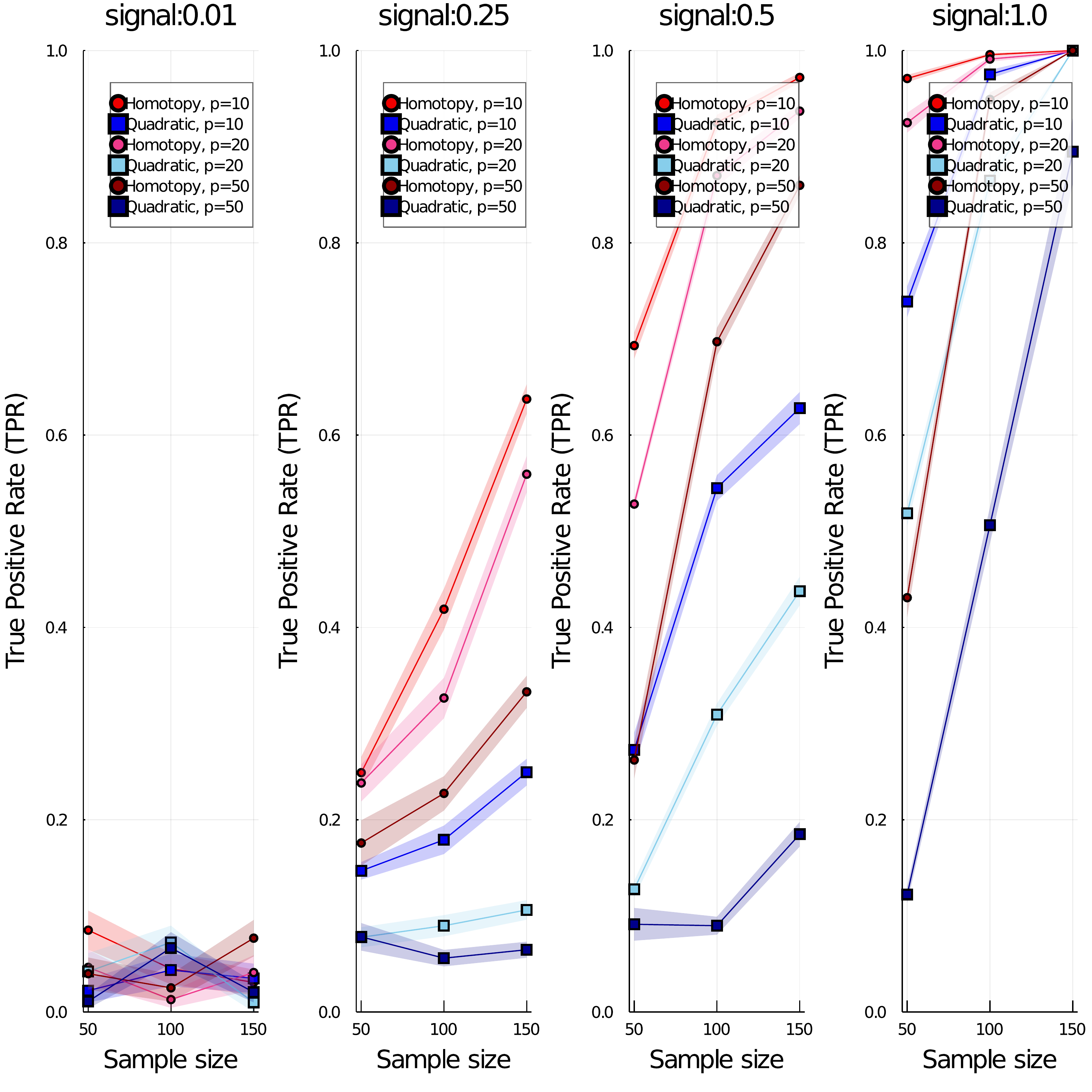}\\
  {\small (b) True Positive Rates (TPRs)} \\
  \includegraphics[width=0.50\linewidth]{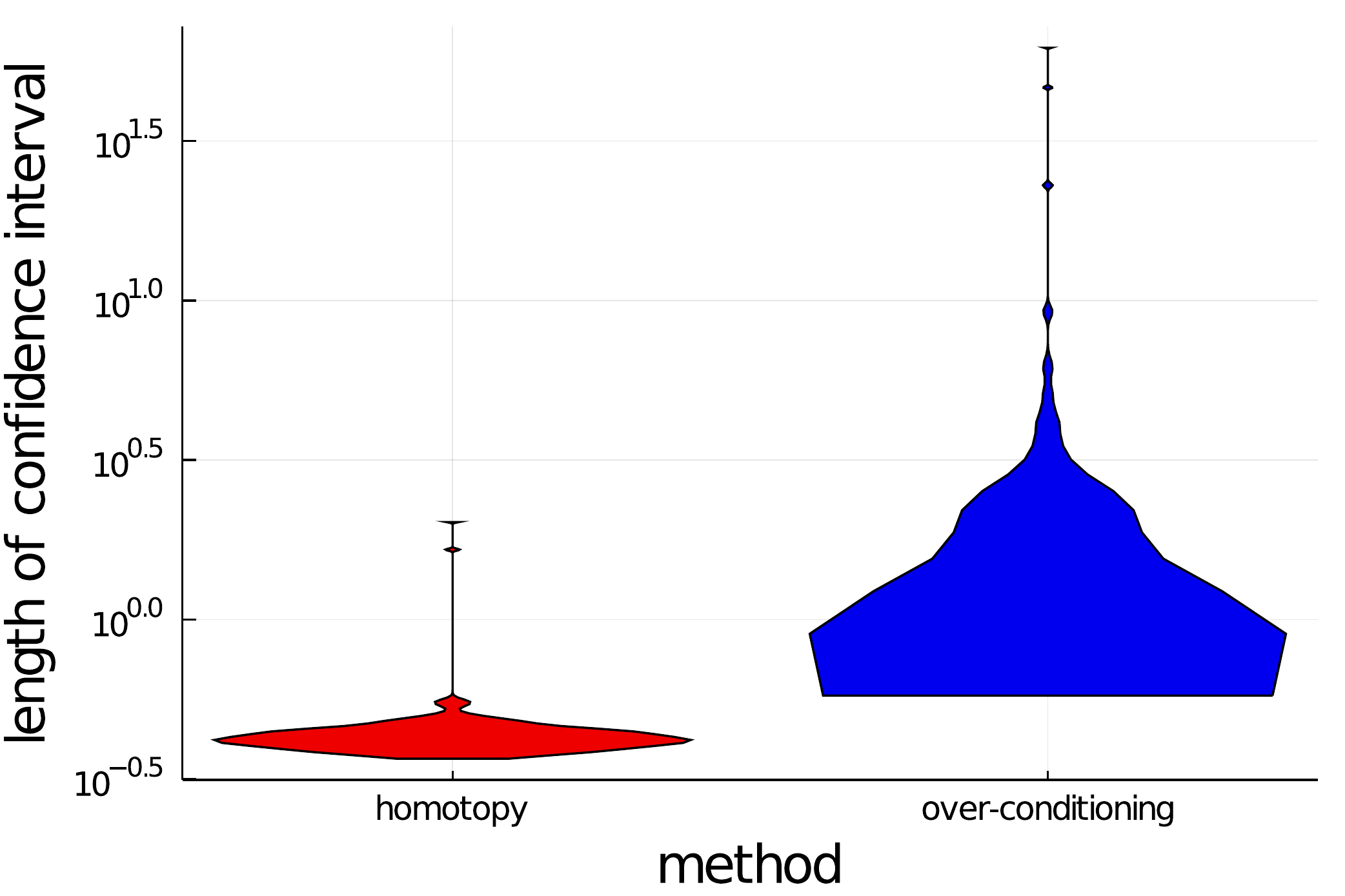}\\
  {\small (c) Confidence Interval (CI) Lengths} \\
  \caption{Results of FB-SFS on synthetic data.}
  \label{fig:exp_fb_sfs_synthetic}  
 \end{center}
\end{figure}

The results of FPRs, TPRs and CIs are shown in Fig. \ref{fig:exp_fb_sfs_synthetic}(a), (b) and (c), respectively.
The FPRs and TPRs were estimated by 100 trials, and the plots in Fig. \ref{fig:exp_fb_sfs_synthetic} (a) and (b) are the averages when the 100 trials were repeated 20 times for FPR and 10 times for TPR, whereas the plots in Fig. \ref{fig:exp_fb_sfs_synthetic}(c) are the results of 100 trials. 
In both methods, the FPRs are properly controlled under the significance level $\alpha$.
Regarding the TPR comparison, it is obvious that {\tt Homotopy} method has the higher power than {\tt Quadratic} because it is minimally conditioned. 
The results of the CIs are consistent with the results of TPRs. 

\paragraph{Real data experiments}
We compared the proposed method ({\tt Homotopy}) and conventional method ({\tt Quadratic}) on the same three real datasets.
From each of the original dataset, we randomly generated sub-sampled datasets with sizes 25, 50 and 100. 
We applied the proposed and the conventional methods to each sub-sampled dataset, examined the cases where the selective $p$-values of a selected feature was different between the two methods, and computed the percentage of times when the proposed method provided smaller selective $p$-values than the conventional one. 
Table \ref{tab:exp_fb_real} shows the results from 1000 trials. 
The proposed method provides smaller selective $p$-values than the conventional method significantly more frequently especially when $n$ is large.

\section{Conclusion}
In this paper, we proposed a more powerful and general conditional SI method for SFS algorithm.
We resolved the over-conditioning problem in existing approaches by introducing the homotopy continuation approach.
The experimental results indicated that the proposed homotopy-based approach is more powerful and general.

\subsection*{Acknowledgement}
This work was partially supported by MEXT KAKENHI (20H00601, 16H06538), JST CREST (JPMJCR1502), RIKEN Center for Advanced Intelligence Project, and RIKEN Junior Research Associate Program.

\appendix

\section{Proofs}
\label{app:proofs}

\subsection{Proof of Lemma 1}
\begin{proof}
Let $(\cH(z), \cS(z)) = \cA_2(\bm y(z))$ for all $z \in \RR$ and write 
$\cH(z) = (\cM_1(z), \ldots, \cM_K(z))$
and
$\cS(z) = (\cS_1(z), \ldots, \cS_K(z))$.
 By conditioning on the histories
 $\cH(z) = \cH^\prime$, 
 \begin{align*}
  \cM_1(z) = \cM_1^\prime, \ldots, \cM_K(z) = \cM_K^\prime 
 \end{align*}
 for all 
 $z \in \RR$
 such that
 $\cH(z) = \cH^\prime$. 
 This indicates that
 \begin{align}
  \label{eq:lemm1_b}
  \left|
  \bm x_{j^\prime_k}^\top \bm r(\bm y(z), X_{\cM^\prime_{k-1}})
  \right|
  \ge
  \pm \bm x_{j}^\top \bm r(\bm y(z), X_{\cM^\prime_{k-1}})
 \end{align} 
 for all 
 $(k, j) \in [K] \times ([p] \setminus \cM^\prime_{k-1})$
 and 
 $z \in \RR$ such that $\cH(z) = \cH^\prime$, where $(j^\prime_1, \ldots, j^\prime_K)$ is the sequence of the selected features when the $K$-step SFS algorithm is applied to the response vector $\bm y^\prime$. 
 By further conditioning on the signs $\cS(z) = \cS^\prime$, \eq{eq:lemm1_b} is written as 
 \begin{align*}
 \cS^\prime_{k} \bm x_{j^\prime_k}^\top \bm r(\bm y(z), X_{\cM^\prime_{k-1}}) \ge \pm \bm x_{j}^\top \bm r(\bm y(z), X_{\cM^\prime_{k-1}})
 \end{align*} 
 for all 
 $(k, j) \in [K] \times ([p] \setminus \cM^\prime_{k-1})$
 and 
 $z \in \RR$ such that $\cH(z) = \cH^\prime$ and $\cS(z) = \cS^\prime$.
 By restricting on a line $\bm y(z) = \bm a + \bm b z, z \in \RR$, the range of $z$ is written as 
 \begin{align}
  \max_{\substack{ k \in [K], \\ j \in [p] \setminus \cM^\prime_{k-1}, \\ d_{(k,j)} > 0} } \frac{e_{(k,j)}}{d_{(k,j)}} \le z \le \min_{\substack{k \in [K], \\ j \in [p] \setminus \cM^\prime_{k-1}, \\ d_{(k,j)} < 0}} \frac{e_{(k,j)}}{d_{(k,j)}}.
 \end{align}
\end{proof}

\subsection{The proof of Lemma 2}
\begin{proof}
 For a set of features $\cM \subseteq [p]$ and a response vector $\bm y(z) = \bm a + \bm b z, z \in \RR$ in a line, let us denote the AIC as a function of $\cM$ and $z$ as ${\rm AIC}(\cM, z)$.
 Subsequently, by substituting $\bm y = \bm a + \bm b z$ into \eq{eq:aic1}, it is written as a quadratic function of $z$ as 
 \begin{align}
  \nonumber
  {\rm AIC}(\cM, z)
  &
  = 
  (\bm b^\top A_{\cM} \bm b) z^2 + 2 (\bm a^\top A_{\cM} \bm b) z + (\bm a^\top A_{\cM} \bm a) 
  \\
  \label{eq:aic_quadratic}
  &+
  2 |\cM|.  
 \end{align}
 Equation \eq{eq:aic_quadratic} represents the range of $z \in \RR$ such that when $\bm y(z)$ is fed into the algorithm, the same history $\cH = \cA_2(\bm y^\prime)$ is obtained.
 Let the sequence of the selected models corresponding to the history $\cH^\prime = \psi(\bm y^\prime)$ be 
 \begin{align*}
  \cH^\prime = (\cM^\prime_1, \ldots, \cM^\prime_K),
 \end{align*}
 where $K$ is the number of steps in the history $\cH$. 
 Next, the event of the history can be fully characterized by comparing AICs as follows:
 \begin{align}
  \nonumber
  &
  {\rm AIC}(\cM^\prime_k, z) \le {\rm AIC}(\cM^\prime_{k-1} \cup \{j\}, z) ~ \forall j \in [p] \setminus \cM_{k-1},
  \\
  \nonumber
  &
  {\rm AIC}(\cM^\prime_k, z) \le {\rm AIC}(\cM^\prime_{k-1} \setminus \{j\}, z) ~ \forall j \in \cM_{k-1},
  \\
  \label{eq:aic_comp1}
  &
  {\rm AIC}(\cM^\prime_k, z) \le {\rm AIC}(\cM^\prime_{k-1}, z),
 \end{align}
 for $k = 1, 2, \ldots, K$
 and
 \begin{align}
  \nonumber
  &
  {\rm AIC}(\cM^\prime_K, z) < {\rm AIC}(\cM^\prime_K \cup \{j\}, z) ~ \forall j \in [p] \setminus \cM_K, 
  \\
  \label{eq:aic_comp2}
  &
  {\rm AIC}(\cM^\prime_K, z) < {\rm AIC}(\cM^\prime_K \setminus \{j\}, z) ~ \forall j \in \cM_K.
 \end{align}
 Here, the first and second inequalities in \eq{eq:aic_comp1} indicate that the selected model at step $k$ has the smallest AIC among all possible choices, the third inequality in \eq{eq:aic_comp1} indicates that the AIC of the selected model at step $k$ is smaller than that at the previous step, and two inequalities \eq{eq:aic_comp2} indicate that the AIC of the  selected model at the final step $K$ cannot be decreased anymore. 
 Because the AIC is written as a quadratic function of $z$ as in \eq{eq:aic_quadratic} under the fixed history $\cH^\prime$, all these conditions are written as quadratic inequalities of $z$. 
 This means that the range of $z \in \RR$ that satisfies these conditions is represented by a finite set of quadratic inequalities of $z \in \RR$. 
\end{proof}


\section{Details of Experiments}
\label{app:exp_setup}

We executed the experiments on Intel(R) Xeon(R) CPU E5-2687W v3 @ 3.10GHz.

\paragraph{Comparison methods in the case of Forward SFS.}
In the case of forward SFS, we remind that $\cA(\bm y)$ results a set of selected featured $\cM$ when applying forward SFS algorithm $\cA$ to $\bm y$.
With a slight abuse of notations, let $\cH(\bm y)$ and $\cS(\bm y)$ respectively denote the history and signs obtained when applying algorithm $\cA$ to $\bm y$.
We compared the following five methods: 
\begin{itemize}
 \item {\tt Homotopy}: conditioning on the selected features $\cM$ (minimal conditioning), i.e., 
 \begin{align*}
 	\bm \eta^T \bm Y \mid \{\cA(\bm Y) = \cA(\bm y), \bm q(\bm Y) = \bm q (\bm y)\}.
 \end{align*}
 
 \item {\tt Homotopy-H}: additionally conditioning on the history $\cH$, i.e., 
 \begin{align*}
 	\bm \eta^T \bm Y \mid \{\cH(\bm Y) = \cH(\bm y), \bm q(\bm Y) = \bm q (\bm y)\}.
 \end{align*}
 Here, we note that  $\cH(\bm Y) = \cH(\bm y)$ already includes the event $\cA(\bm Y) = \cA(\bm y)$.
 
 \item {\tt Homotopy-S}: additionally conditioning on the signs, i.e., 
 \begin{align*}
 	\bm \eta^T \bm Y \mid \{\cA(\bm Y) = \cA(\bm y), \cS(\bm Y) = \cS(\bm y), \bm q(\bm Y) = \bm q (\bm y)\}.
 \end{align*}
 
 \item {\tt Polytope} \citep{tibshirani2016exact}: additionally conditioning on both history $\cH$ and signs $\cS$, i.e., 
 \begin{align*}
 	\bm \eta^T \bm Y \mid \{\cH(\bm Y) = \cH(\bm y), \cS(\bm Y) = \cS(\bm y), \bm q(\bm Y) = \bm q (\bm y)\}.
 \end{align*}
 This definition is equivalent to 
  \begin{align*}
 	\bm \eta^T \bm Y \mid \{\cA_2(\bm Y) = \cA_2(\bm y), \bm q(\bm Y) = \bm q (\bm y)\},
 \end{align*}
 where $\cA_2(\cdot)$ is defined in \S3.
 
 \item {\tt DS}: data splitting is the commonly used procedure for the purpose of selection bias correction. 
In this approach, the data is randomly divided in two halves — first half used for model selection and the other for inference.
\end{itemize}

\paragraph{Comparison methods in the case of Forward-Backward SFS.}
In the case of forward-backward SFS,  $\cA(\bm y)$ results a set of selected featured $\cM$ when applying forward-backward SFS algorithm $\cA$ to $\bm y$,
and $\cA_2(\bm y)$ results the history.
We compared the following two methods: 
\begin{itemize}
 \item {\tt Homotopy}: conditioning on the selected features $\cM$ (minimal conditioning), i.e., 
 \begin{align*}
 	\bm \eta^T \bm Y \mid \{\cA(\bm Y) = \cA(\bm y), \bm q(\bm Y) = \bm q (\bm y)\}.
 \end{align*}
 \item {\tt Quadratic}: additionally conditioning on the history $\cH$ (implemented by using quadratic inequality-based conditional SI in \citep{loftus2015selective}), i.e., 
 \begin{align*}
 	\bm \eta^T \bm Y \mid \{\cA_2(\bm Y) = \cA_2(\bm y), \bm q(\bm Y) = \bm q (\bm y)\}.
 \end{align*}
\end{itemize}

\paragraph{Definition of TPR.} In SI, we only conduct statistical testing when there is at least one hypothesis discovered by the algorithm.
Therefore, the definition of TPR, which can be also called \emph{conditional power}, is as follows:
\begin{equation*}
	{\rm TPR} = \frac{{\rm \#\ detected\ \&\ rejected}}{{\rm \#\ detected}},
\end{equation*}
where ${\rm \#\ detected}$ is the number of truly positive features selected by the algorithm (e.g., SFS) and  ${\rm \#\ rejected}$ is the number of truly positive features whose null hypothesis is rejected by SI.

\begin{figure}[p]
\centering
\includegraphics[width=.5\linewidth]{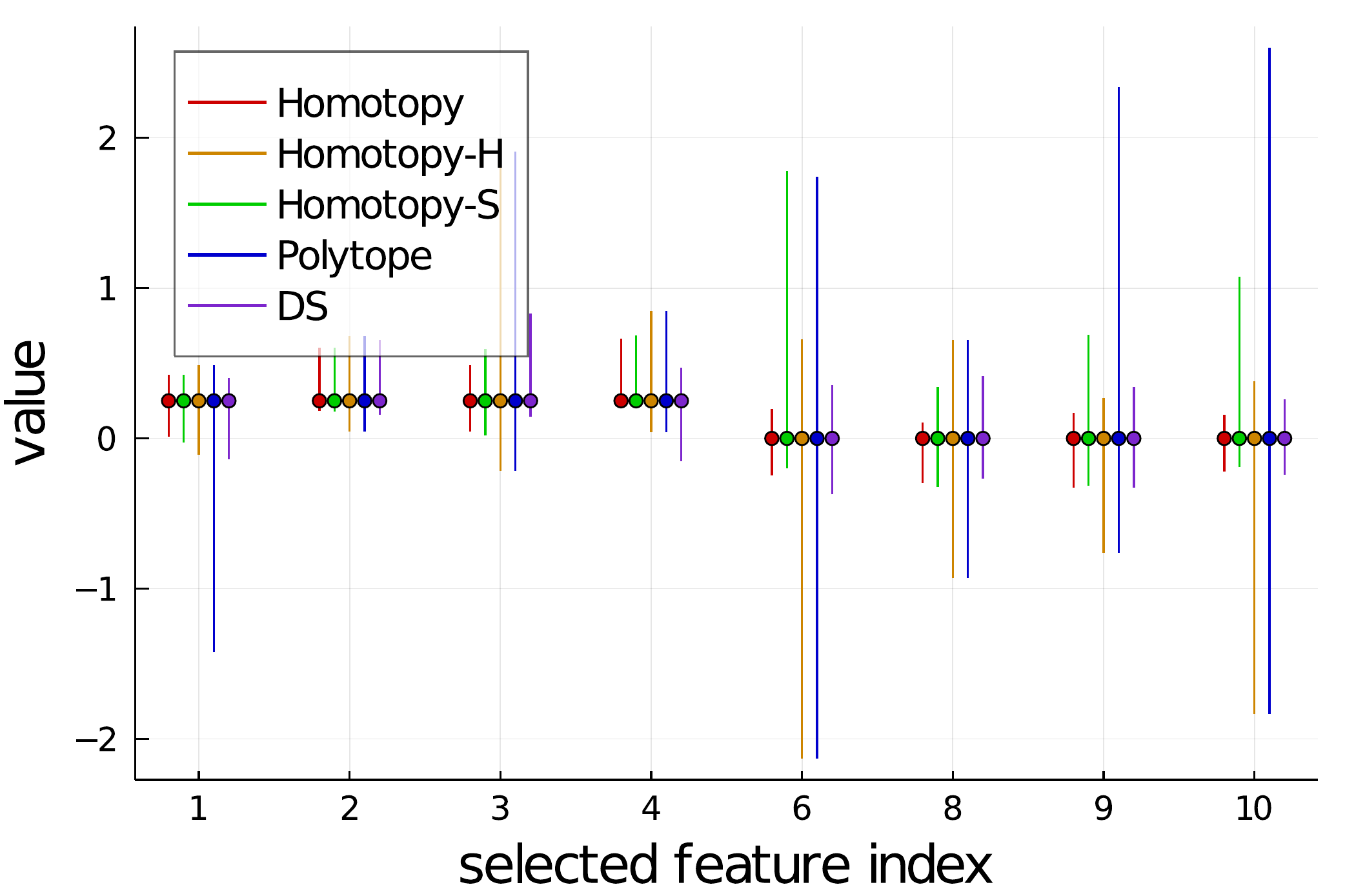}
\caption{Demonstration of confidence interval.}
\label{fig:ci_demo}
\end{figure}

\paragraph{Demonstration of confidence interval (forward SFS).}
We generated $n=100$ outcomes as $y_{i} = \bm{x}_{i}^{\top}\bm{\beta} + \varepsilon_{i}, ~i = 1, ..., n$, where $\bm{x}_{i} \sim \mathcal{N}(0, I_{p})$ and $\varepsilon_{i} \sim \mathcal{N}(0, 1)$.
We set $p=10$, $K=9$ and $\bm{\beta} = \left[ 0.25, 0.25, 0.25, 0.25, 0.25, 0, 0, 0, 0, 0 \right]^{\top}$.
We note that the number of selected features between the four options of conditional SI methods ({\tt Homotopy}, {\tt Homotopy-H}, {\tt Homotopy-S}, {\tt Polytope}) and DS can be different. Therefore, for a fair comparison, we only consider the features that are selected in both cases.
Figure \ref{fig:ci_demo} shows the demonstration of CIs for each selected feature.
The results are consistent with Fig. \ref{fig:exp_f_sfs_synthetic} (b).

\section{Experiments on Computational Aspects (Forward SFS)}
\label{app:exp_computational}

\begin{figure}[p]
\centering
\includegraphics[width=.5\linewidth]{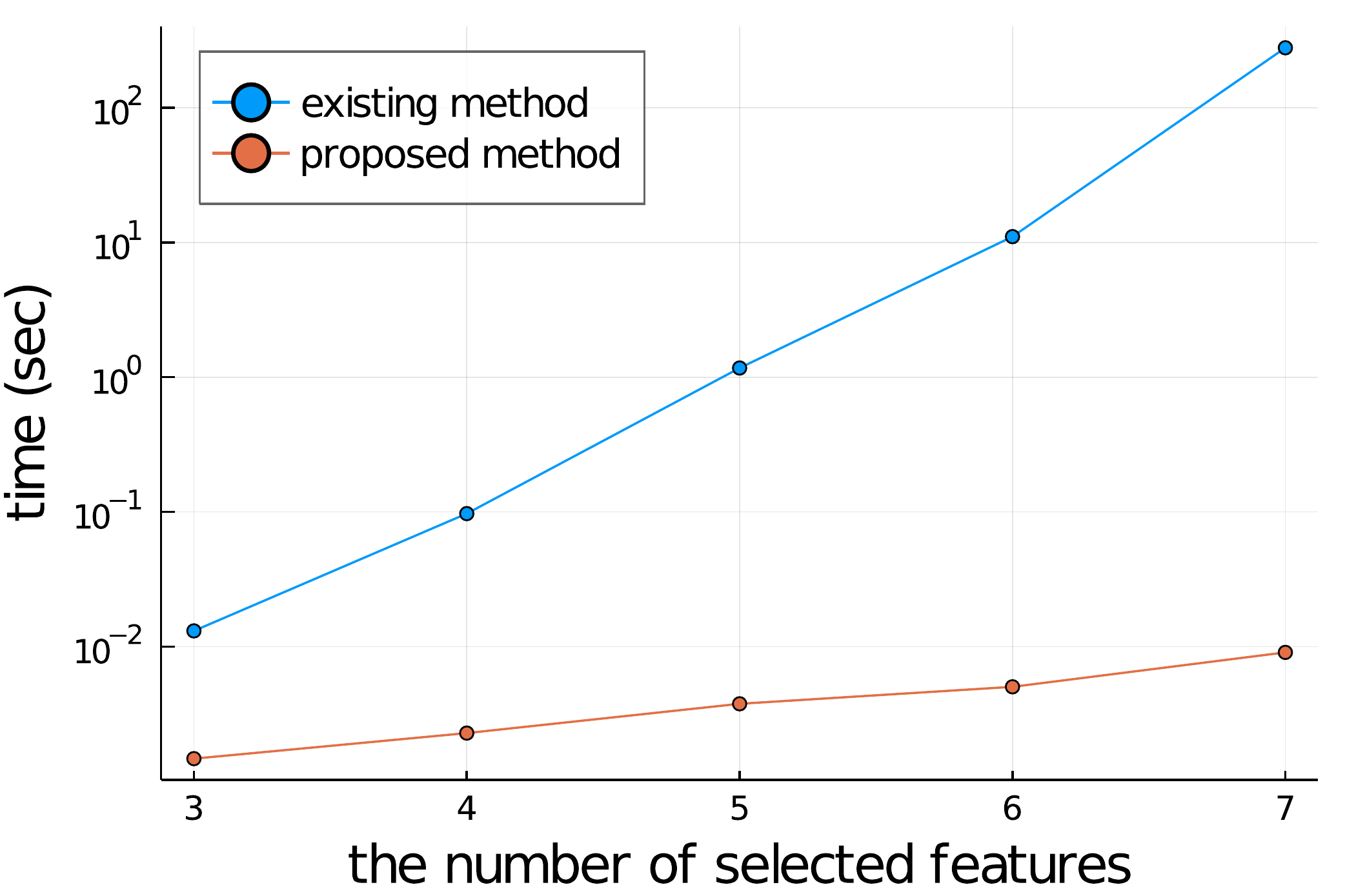}
\caption{The result of comparing the computational time between the proposed method and the existing method with an $n=50, ~p=10$ artificial dataset.}
\label{fig:vs_naive}
\end{figure}

We demonstrate the computational efficiency of the proposed {\tt Homotopy} method.
We generated $n$ outcomes as $y_{i} = \bm{x}_{i}^{\top}\bm{\beta} + \varepsilon_{i}, ~i = 1, ..., n$, where $\bm{x}_{i} \sim \mathcal{N}(0, I_{p})$ and $\varepsilon_{i} \sim \mathcal{N}(0, 1)$.
We set $n=50$ and $p=10$.
In Figure \ref{fig:vs_naive}, we show the results of comparing the computational time between the proposed {\tt Homotopy} method and the existing method.
For the existing study, if we want to keep high statistical power, we have to enumerate a huge number of all the combinations of histories and signs
$
2^{K} \times K!, 
$
which is only feasible when the number of selected features is fairly small.
We observe in blue plots in Fig. \ref{fig:vs_naive} that the computational cost of existing method is exponentially increasing with the number of selected features.
With the proposed method, we are able to significantly reduce the computational cost while keeping high power.

\begin{figure}[p]
\centering
\includegraphics[width=.5\linewidth]{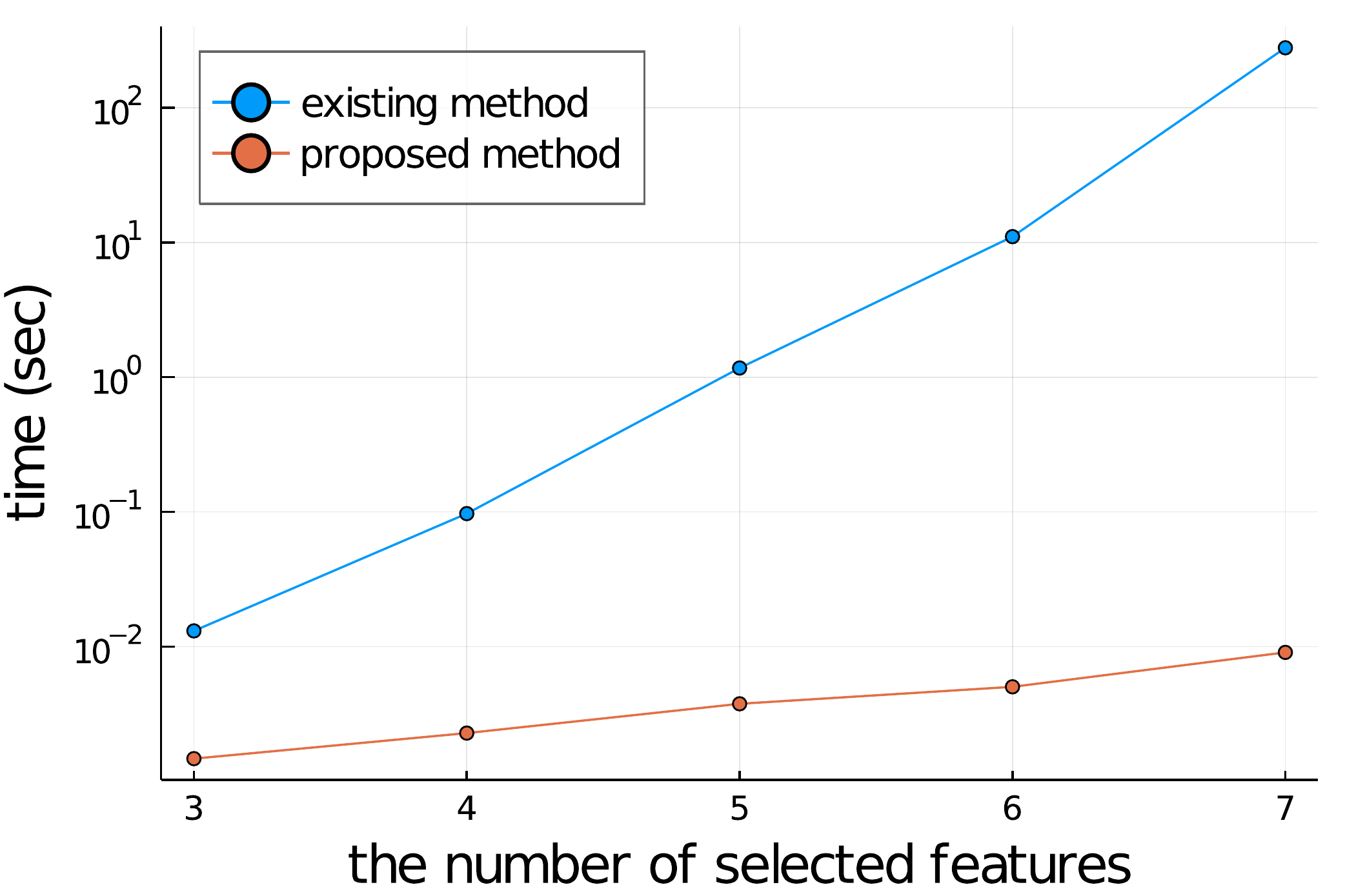}
\caption{The number of polytopes intersecting the line $z$ that we need to consider. The solid lines are shown the sample averages.}
\label{fig:polytope_num}
\end{figure}

One might wonder how we can circumvent the computational bottleneck of exponentially increasing number of polytopes. 
Our experience suggests that, by focusing on the line along the test-statistic in data space, we can skip majority of the polytopes that do not affect the truncated Normal sampling distribution because they do not intersect with this line. 
In other words, we can skip majority of combinations of histories and signs that never appear.
We show the violin plot of the actual numbers of intervals of the test statistic $z$ that involves in the construction of truncated sampling distribution in Figure \ref{fig:polytope_num}. 
Here, we set $n=250$ and $=50$.
Regarding {\tt Homotopy} and {\tt Homotopy-S}, the number of polytopes linearly increases when increasing $K$.
This is the reason why our method is highly efficient.
In regard to {\tt Homotopy-H}, the number of polytopes decreases because the \emph{history} constraint is too strict
when $K$ is increased.

\section{Experiments on Robustness (Forward-Backward SFS)}
\label{app:exp_robustness}
\begin{figure*}[p]
\begin{subfigure}{.5\textwidth}
  \centering

 \includegraphics[width=0.81\linewidth]{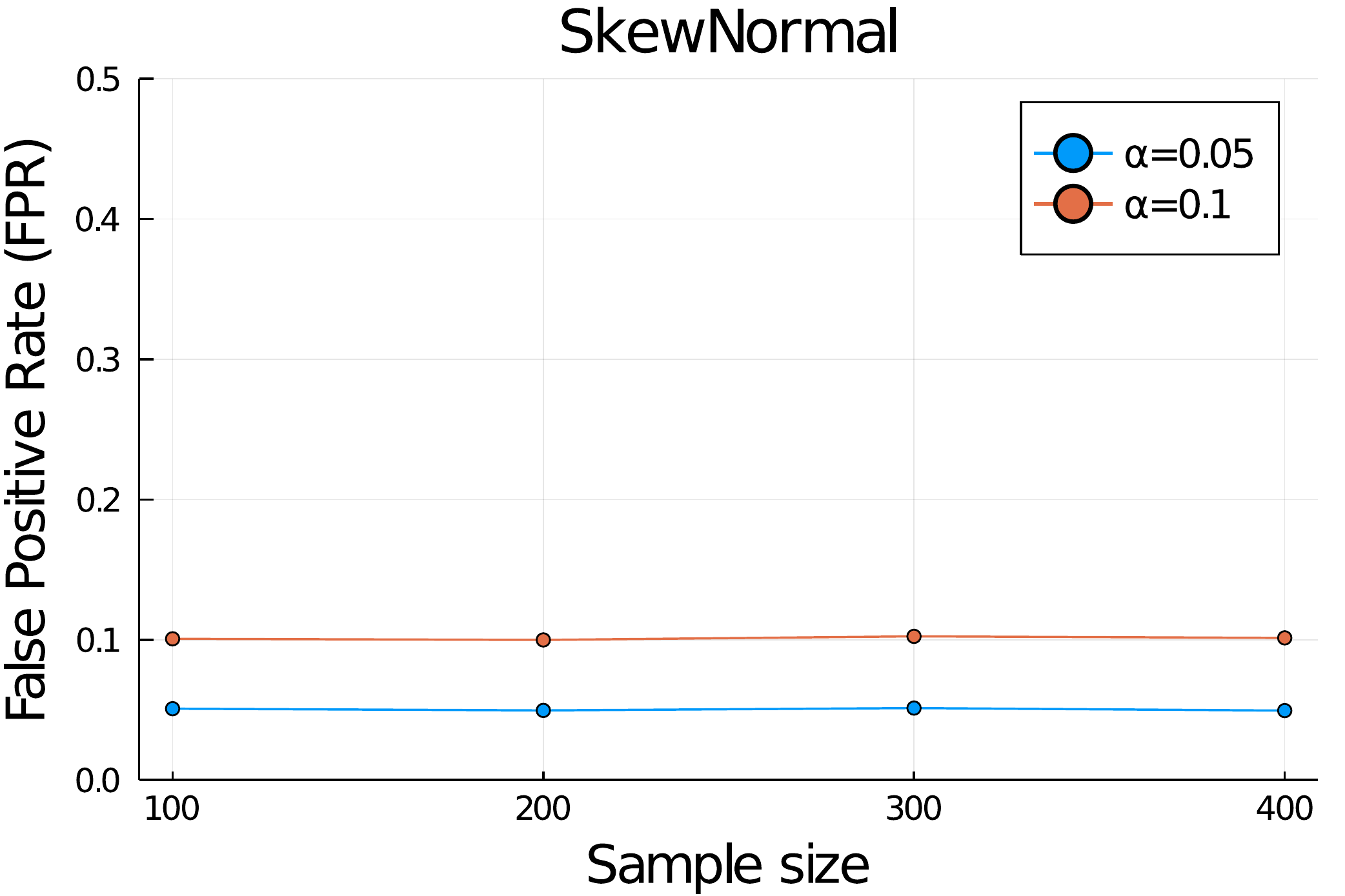}  
\end{subfigure}
\begin{subfigure}{.5\textwidth}
  \centering
  \includegraphics[width=.81\linewidth]{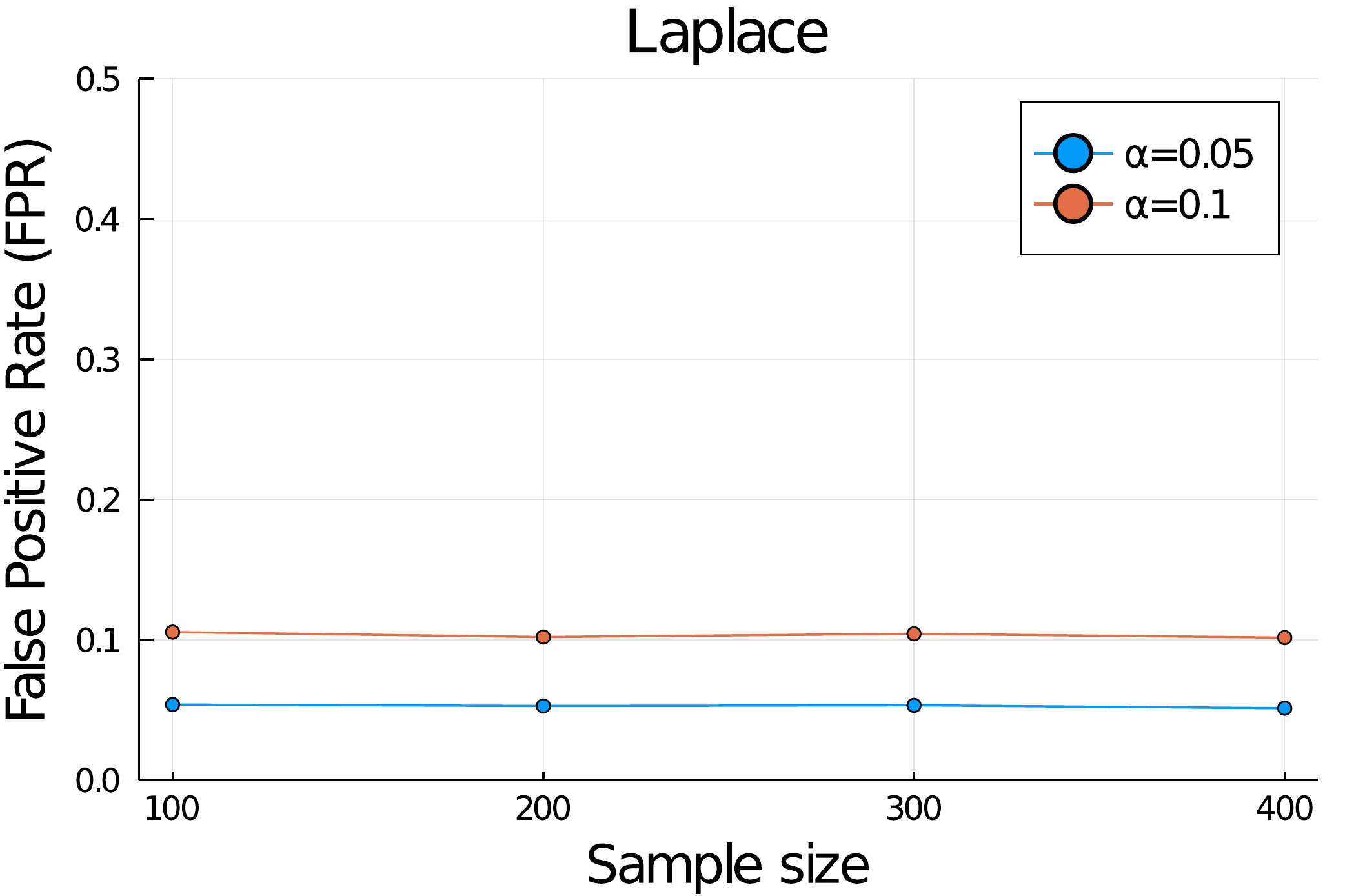}  
\end{subfigure}
\begin{subfigure}{.5\textwidth}
  \centering
  \includegraphics[width=.81\linewidth]{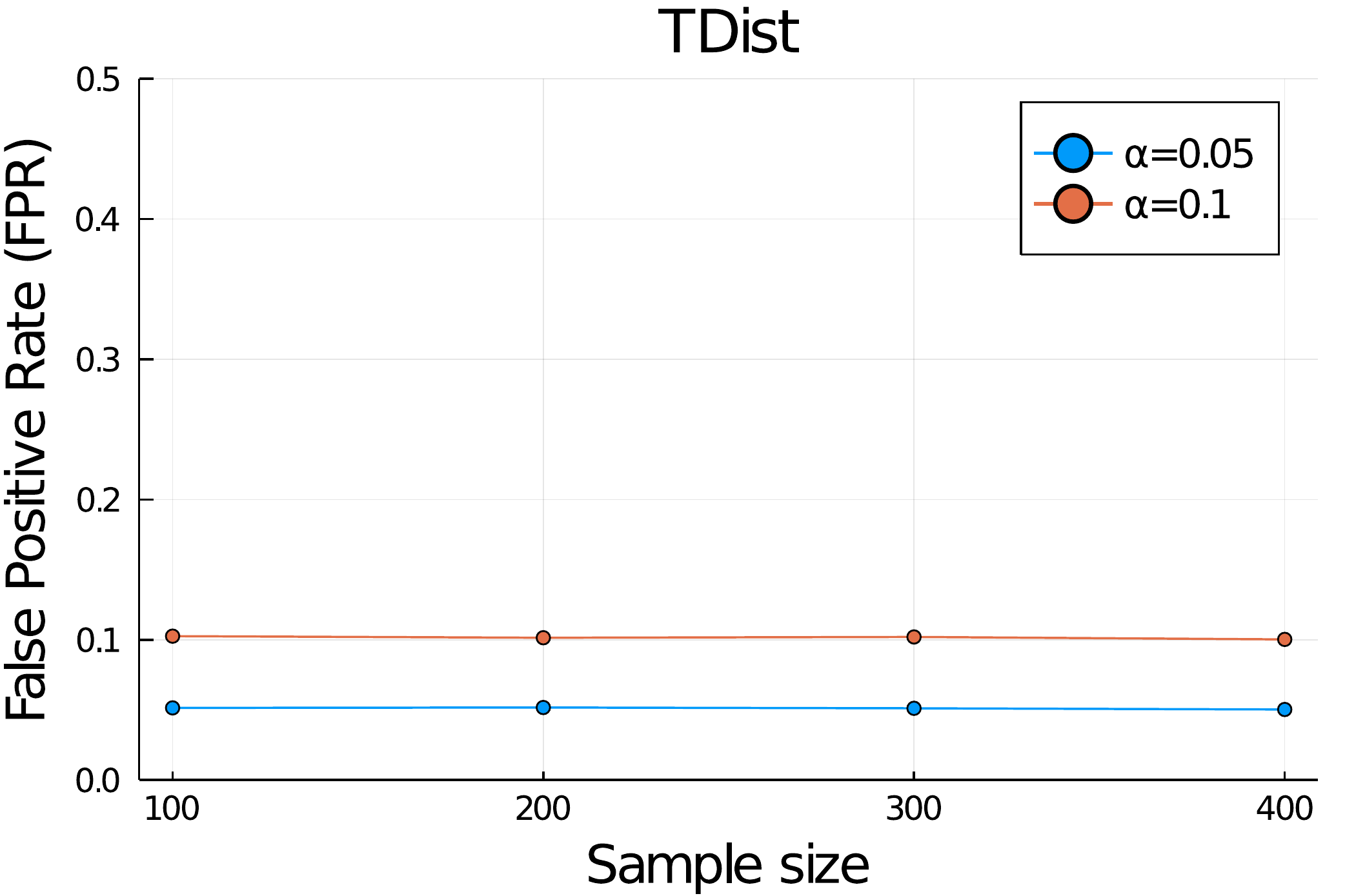}  
\end{subfigure}
\begin{subfigure}{.5\textwidth}
  \centering
  \includegraphics[width=.81\linewidth]{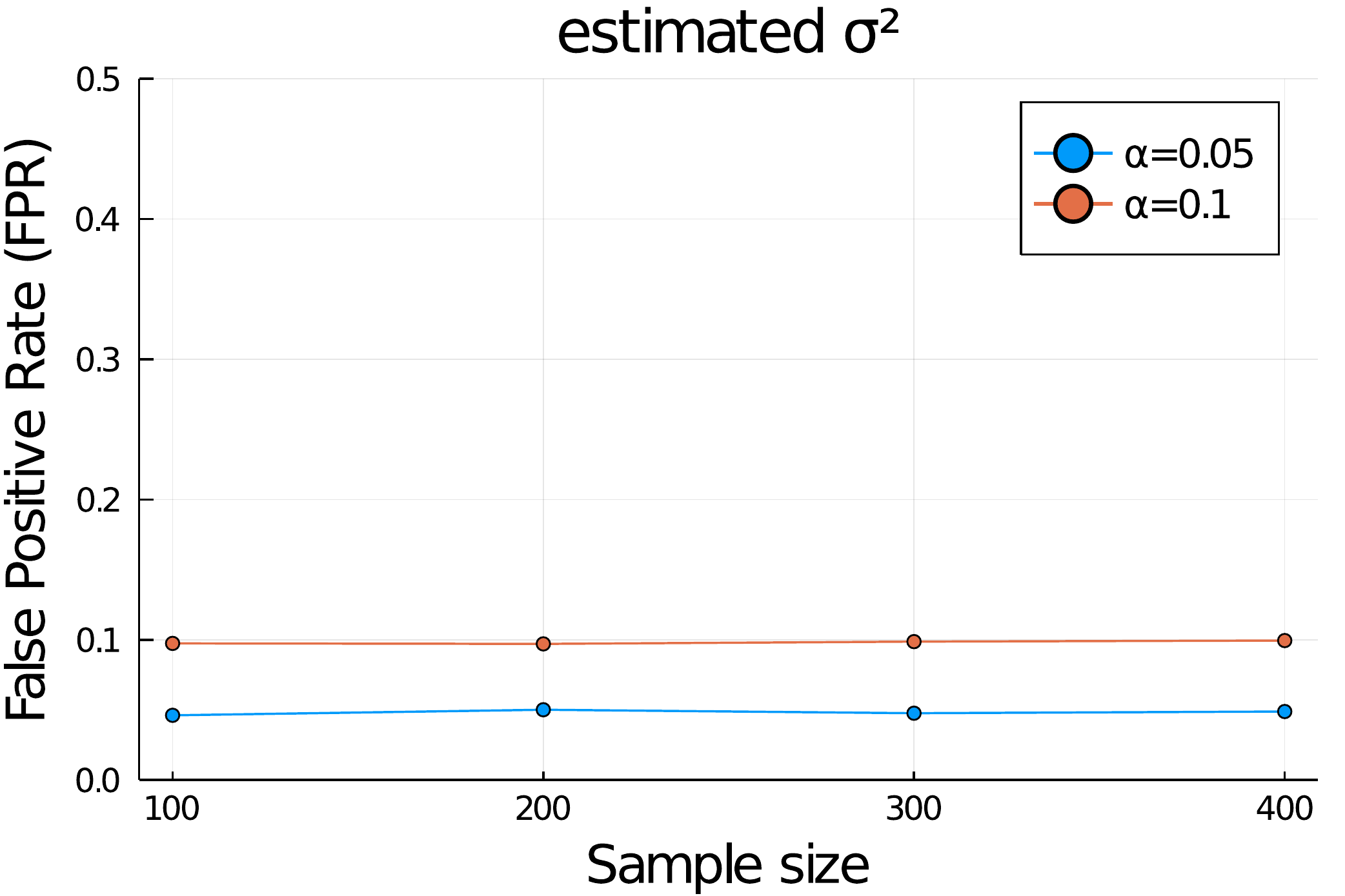}  
\end{subfigure}
\caption{The robustness of the proposed method in terms of the FPR control.}
\label{fig:fig_robustness}
\end{figure*}

We applied our proposed method to the cases where the noise follows Laplace distribution, skew normal distribution (skewness coefficient 10), and $t_{20}$ distribution. 
We also conducted experiments when $\sigma^2$ was also estimated from the data.
We generated $n$ outcomes as $y_i = \bm x_i^\top \bm \beta + \veps_i$, 
$i = 1, ..., n$, 
where 
$p = 5, \bm x_i \sim \NN(0, I_p)$, 
and $\veps_i$ follows Laplace distribution, skew normal distribution, or $t_{20}$ distribution with zero mean and standard deviation was set to 1.
In the case of estimated $\sigma^2$, $\veps_i \sim \NN(0, 1)$.
We set all elements of $\bm \beta$ to 0, and set $\lambda = 0.5$.
For each case, we ran 1,200 trials for each $n \in \{100, 200, 300, 400\}$. 
The FPR results are shown in Figure \ref{fig:fig_robustness}.
Although we only demonstrate the results for the case of forward-backward SFS algorithm, the extension to forward SFS algorithm with similar setting is straightforward.

\section{Homotopy-based SI for Cross-Validation (Forward SFS)}
\label{app:cv}
In this section, we introduce a method for SI conditional also on the selection of the number of selected features $K$ via cross-validation.
Consider selecting the number of steps $K$ in the SFS method from a given set of candidates $\cK = \{K_1, \ldots, K_L\}$ where $L$ is the number of candidates.
When conducting cross-validation on the observed dataset $(X, \bm y)$, suppose that $\cV(\bm y) = K_{\rm selected} \in \cK$ is the event that $K_{\rm selected}$ is selected as the best one.
The test-statistic for the selected feature $j$ when applying the SFS method with $K_{\rm selected}$ steps to $(X, \bm y)$ is then defined as 
\begin{align}\label{eq:_conditional_inference_cv}
 \bm \eta^\top \bm Y \mid 
 &
 \{ \cA(\bm Y) = \cA(\bm y),
 \cV(\bm Y) = K_{\rm selected}, \bm q(\bm Y) = \bm q(\bm y) \}.
\end{align}
We note that $\cA(\cdot)$ and $\bm q (\cdot)$ depend on $K_{\rm selected}$ but we omit the dependence for notational simplicity.
The conditional data space in (\ref{eq:parametrized_data_space}) with the event of selecting $K_{\rm selected}$ is then written as 
\begin{align}
 \cY =\{\bm y(z) = \bm a + \bm bz \mid z \in \cZ_{\rm CV}\},
\end{align}
where $\cZ_{\rm CV} = \{z \in \RR \mid \cA(\bm y(z)) = \cA(\bm y), \cV(\bm y(z)) = K_{\rm selected}\}$.
The truncation region $\cZ_{\rm CV}$ can be obtained by the intersection of the following two sets: 
\begin{align*}
\cZ_1
= \{z \in \RR \mid \cA(\bm y(z)) = \cA(\bm y)\}
\quad \text{ and } \quad
 \cZ_2
 = \{z \in \RR \mid \cV(\bm y(z)) = K_{\rm selected}\}.
\end{align*} 
Since the former $\cZ_1$ can be obtained by using the method described in \S 3, the remaining task is to identify the latter $\cZ_2$. 

For notational simplicity, we consider the case where the dataset $(X, \bm y)$ is divided into training and validation sets, and the latter is used for selecting $K_{\rm selected}$. 
The following discussion can be easily extended to cross-validation scenario. 
Let us re-write 
\begin{align*}
 (X, \bm{y}) = \left((X^{\rm tr}\ X^{\rm va} )^\top \in \RR^{n \times p}, (\bm y^{\rm tr}\ \bm y^{\rm va})^\top \in\RR^{n} \right).
\end{align*}
With a slight abuse of notation, for $K \in \cK$, let $\cM_K(\bm y^{\rm tr} (z))$ be the set of selected features by applying $K$-step SFS method to $(X^{\rm tr}, \bm y^{\rm tr}(z))$.
The validation error is then defined as 
\begin{align}
 \label{eq:cv_error}
 E_K(z) = \|\bm y^{\rm va}(z) - X^{\rm va}_{\cM_K(\bm y^{\rm tr}(z))} \hat{\bm \beta}_K(z)\|^2_2,
\end{align}
where $\hat{\bm \beta}_K(z) = \left({X^{\rm tr}_{\cM_K(\bm y^{\rm tr}(z))}}^\top X^{\rm tr}_{\cM_K(\bm y^{\rm tr}(z))} \right)^{-1} {X^{\rm tr}_{\cM_K(\bm y^{\rm tr}(z))}}^\top \bm y^{\rm tr} (z)$.
Then, we can write 
\begin{align*}
 \cZ_2 = \{z \in \RR \mid E_{K_{\rm selected}}(z) \leq E_K(z) \text{ for any } K \in \cK \}.
\end{align*}
Since the validation error $E_K(z)$ in \eq{eq:cv_error} is a picecewise-quadratic function of $z$, we have a corresponding picecewise-quadratic function of $z$ for each $K \in \cK$. 
The truncation region $\cZ_2$ can be identified by the intersection of the intervals of $z$ in which the validation error $E_{K_{\rm selected}}(z)$ corresponding to $K_{\rm selected}$ is minimum among a set of picecewise-quadratic functions for all the other $K \in \cK$.

Loftus (2015) already discussed that it is possible to consider cross-validation event into conditional SI framework.
However, his method is highly over-conditioned in the sense that additional conditioning on all intermediate models in the process of cross-validation is required.
Our method described above is minimumly-conditioned SI in the sense that our inference is conducted based exactly on the conditional sampling distribution of the test-statistic in \eq{eq:_conditional_inference_cv} without any extra conditions. 

\begin{figure}[p]
\centering
\includegraphics[width=0.5\linewidth]{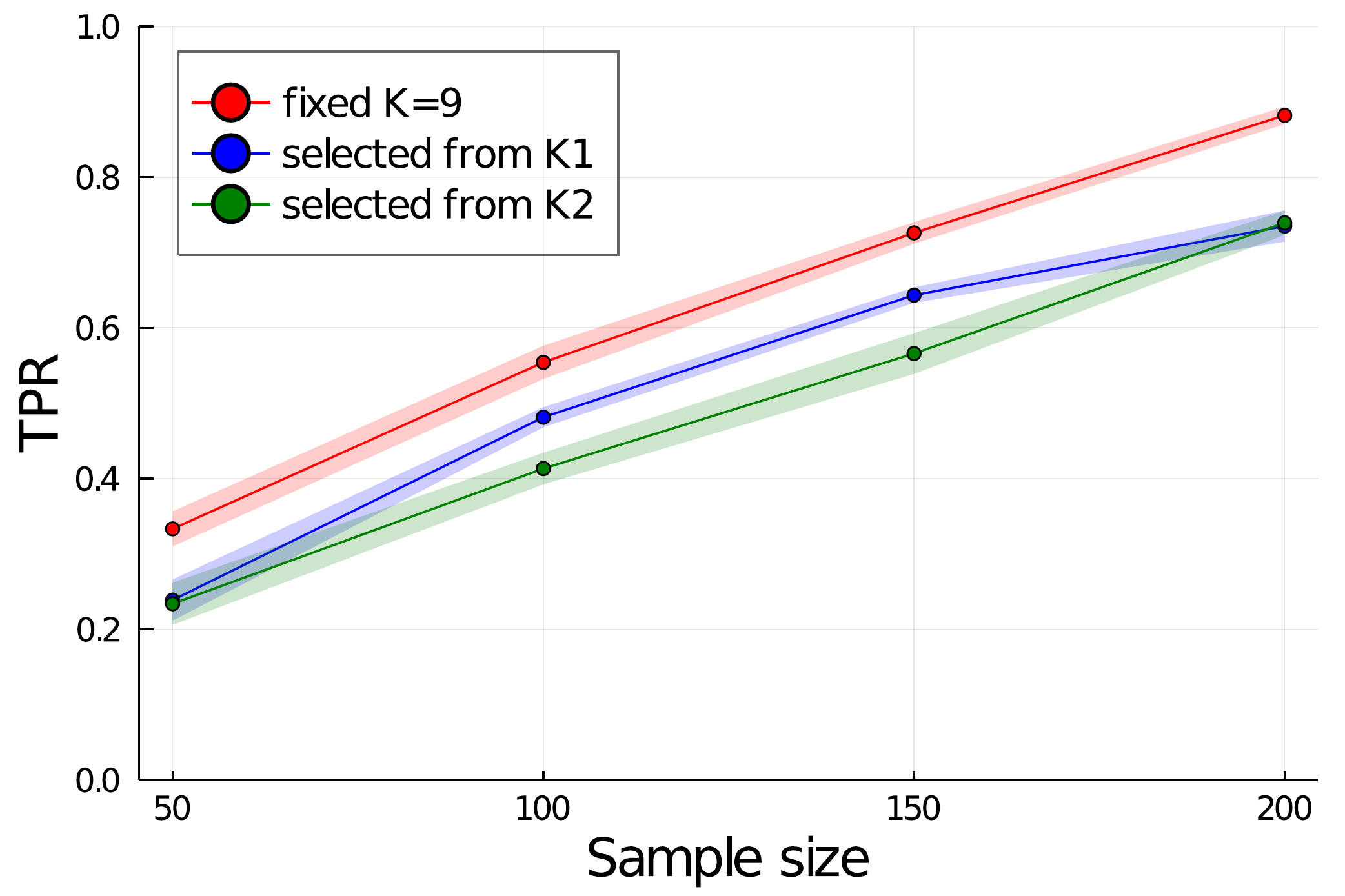}
\caption{Demonstration of TPR when accounting cross-validation selection event.}
\label{fig:TPR_CV}
\end{figure}
\begin{figure}[p]
\centering
\includegraphics[width=0.5\linewidth]{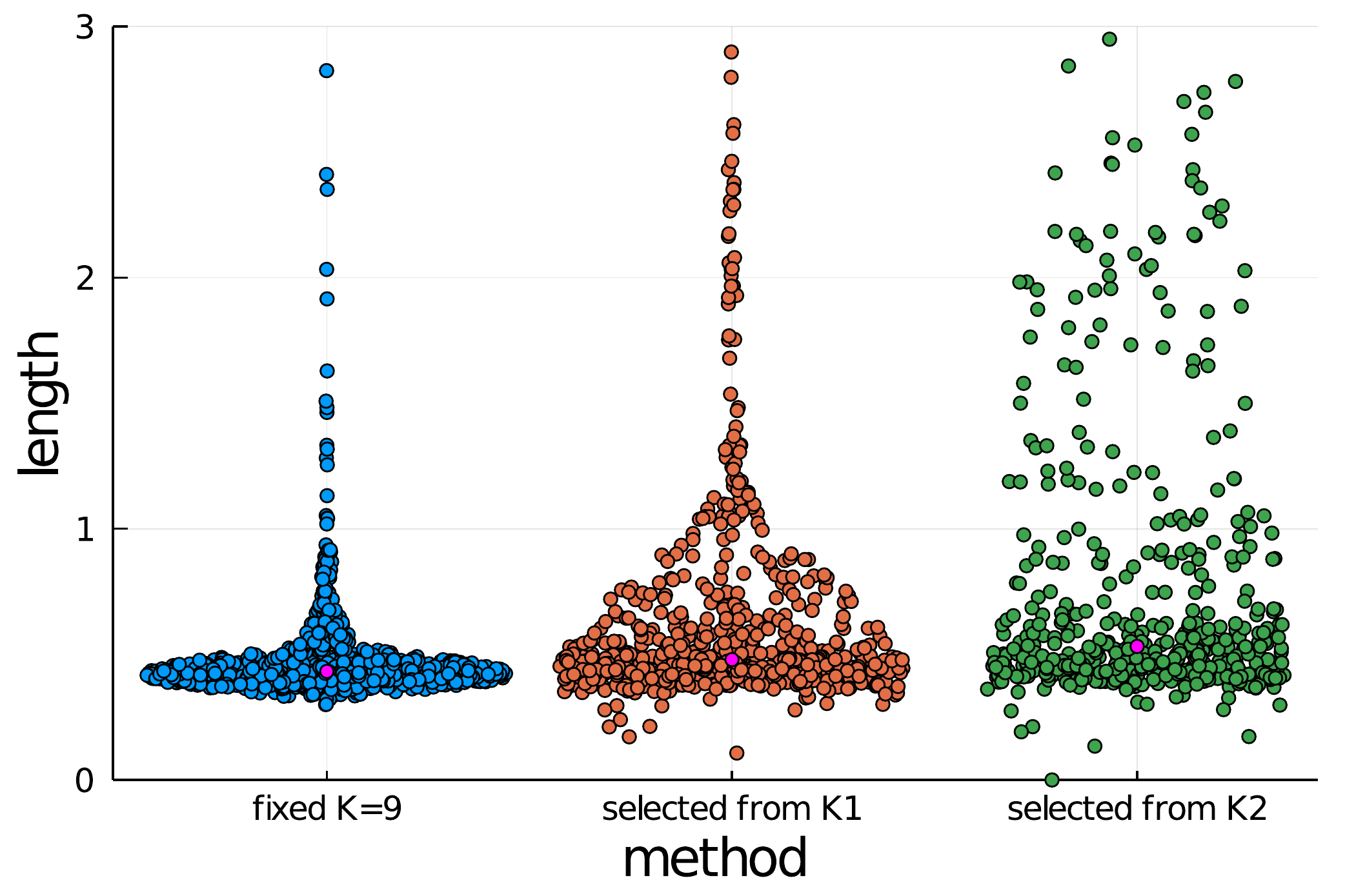}
\caption{Demonstration of CI length when considering cross-validation selection event.}
\label{fig:CI_CV}
\end{figure}

For the experiments on cross-validation, we demonstrate the TPRs and the CIs between the cases when $K=9$ is fixed and $K$ is selected from the set $\cK_{1} = \left\{ 3, 6, 9 \right\}$, or $\cK_{2} = \left\{ 1, 2, \ldots, 10 \right\}$ using $5$-fold cross-validation. 
We set $p=10$, only the first elements of $\bm{\beta}$ was set to $0.25$, and all the rest were set to $0$.
We show that the TPR tends to decrease when increasing the size of $\cK$  in Figure \ref{fig:TPR_CV}.
This is due to the fact that when we increase the size of $\cK$, we have to condition on more information which leads to shorter truncation interval and results low TPR. 
The TPR results are consistent with the CI results shown in Figure \ref{fig:CI_CV}.
In other words, when increasing the size of $\cK$, the lower the TPR is, the longer the length of CI becomes.

\newpage


\end{document}